\documentclass[pdflatex,sn-mathphys-num]{sn-jnl}% Math and Physical Sciences Numbered Reference Style
\usepackage{graphicx}%
\usepackage{multirow}%
\usepackage{amsmath,amssymb,amsfonts}%
\usepackage{amsthm}%
\usepackage{mathrsfs}%
\usepackage[title]{appendix}%
\usepackage{textcomp}%
\usepackage{manyfoot}%
\usepackage{booktabs}%
\usepackage{algorithm}%
\usepackage{algorithmicx}%
\usepackage{algpseudocode}%
\usepackage{listings}%
\usepackage[table,xcdraw]{xcolor}
\usepackage{rotating}
\usepackage{colortbl}
\definecolor{c1d}{RGB}{255, 240, 195} % yellow - DINOv2-Base + NetVLAD
\definecolor{c2d}{RGB}{200, 225, 250} % blue - DINOv2-Base + SALAD
\definecolor{c3d}{RGB}{200, 240, 210} % green - DINOv2-Base + BoQ
\definecolor{c4d}{RGB}{255, 218, 190} % orange - DINOv2-Large + NetVLAD
\definecolor{c5d}{RGB}{230, 215, 250} % purple - DINOv2-Large + SALAD
\definecolor{c6d}{RGB}{255, 215, 235} % pink - DINOv2-Large + BoQ
\theoremstyle{thmstyleone}%
\theoremstyle{thmstyletwo}%

\theoremstyle{thmstylethree}%

\begin{document}

\title[Article Title]{Are All Tokens Necessary for Visual Place Recognition? An Empirical Study of Token Reduction for Efficient Inference}


\author[1,2]{\fnm{Tong} \sur{Jin}}\email{jintong@sia.cn}

\author*[1]{\fnm{Yunpeng} \sur{Liu}}\email{ypliu@sia.cn}

\author[1,2]{\fnm{Shuyu} \sur{Hu}}\email{hushuyu@sia.cn}

\author[1,2]{\fnm{Qinghua} \sur{Zhang}}\email{zhangqinghua@sia.cn}

\author[3]{\fnm{Ruize} \sur{Han}}\email{hanruize@suat-sz.edu.cn}

\author[3]{\fnm{Song} \sur{Wang}}\email{wangsong@suat-sz.edu.cn}

\author*[3]{\fnm{Feng} \sur{Lu}}\email{lufeng@suat-sz.edu.cn}

\affil[1]{\orgname{Shenyang Institute of Automation, Chinese Academy of Sciences}}

\affil[2]{\orgname{University of Chinese Academy of Sciences}}

\affil[3]{\orgname{Shenzhen University of Advanced Technology}}


\abstract{
Recent visual place recognition (VPR) methods based on vision transformers, particularly foundation models, have achieved remarkable recognition performance. However, these models process all visual tokens throughout the entire network, resulting in substantial computational overhead, which hinders their deployment in real-time and resource-constrained scenarios. A natural question thus arises: are all visual tokens necessary for VPR?
To answer this question, we present the first systematic benchmark of token reduction for efficient visual place recognition. Our benchmark comprehensively evaluates representative token pruning, token merging, and hybrid pruning-merging methods across multiple state-of-the-art VPR models and diverse benchmark datasets covering urban, suburban, and natural environments. We further investigate token reduction from multiple perspectives, including recognition performance under different reduction configurations, computational complexity, inference speed, qualitative visualization, and deployment efficiency on edge devices. 
Through extensive experiments and in-depth analysis, our benchmark reveals multiple important characteristics of token reduction in VPR and provides several practical insights into the trade-offs between accuracy and inference efficiency. For example, token reduction can reduce computational cost by up to 29\% and improve throughput by up to 44\%, while incurring less than 1\% degradation in recognition accuracy. Overall, this work establishes a comprehensive foundation for future research on token-efficient VPR and efficient visual retrieval systems.
Our codes and models will be available at \url{https://github.com/Tong-Jin01/TokenReduction4VPR}.
}


\maketitle

\section{Introduction}
\label{sec:introduction}
Visual Place Recognition (VPR) aims to determine the location of a query image by matching it against a database of geo-tagged references \cite{vprsurvey1, vprsurvey2, vprbenchmark}, which serves as a fundamental capability for many computer vision and robotics applications, such as mobile robotics \cite{vpr4mr,vpr4ad}, augmented reality \cite{vpr4ar}, and SLAM \cite{vpr4slam}. Despite its importance, VPR remains challenging due to severe appearance variations caused by changes in viewpoint and long-term environmental dynamics. These challenges are further exacerbated by the operational constraints of practical VPR systems, which are often required to operate in real time under limited resources. Consequently, achieving robust recognition while maintaining efficient inference remains a central challenge.

VPR is commonly formulated as an image retrieval task \cite{unifying}. Given a query image and a geo-tagged database, each image is passed through a backbone network to extract local features, which are then encoded into a global descriptor via an aggregation module. Place recognition is then performed via nearest neighbor search in the feature space to retrieve images that best match the query. The evolution from CNN-based backbones to vision transformers, and more recently to large-scale foundation models such as DINOv2 \cite{dinov2}, has brought substantial accuracy improvements. Yet this trend towards increasingly powerful models inevitably introduces higher computational overhead, especially when employing larger models or higher-resolution inputs \cite{selavpr++,effovpr}. This accuracy-efficiency dilemma has become a key obstacle to deploying advanced VPR models in practical real-time systems.

\begin{figure}[t]
    \centering
    \includegraphics[width=1.0\linewidth]{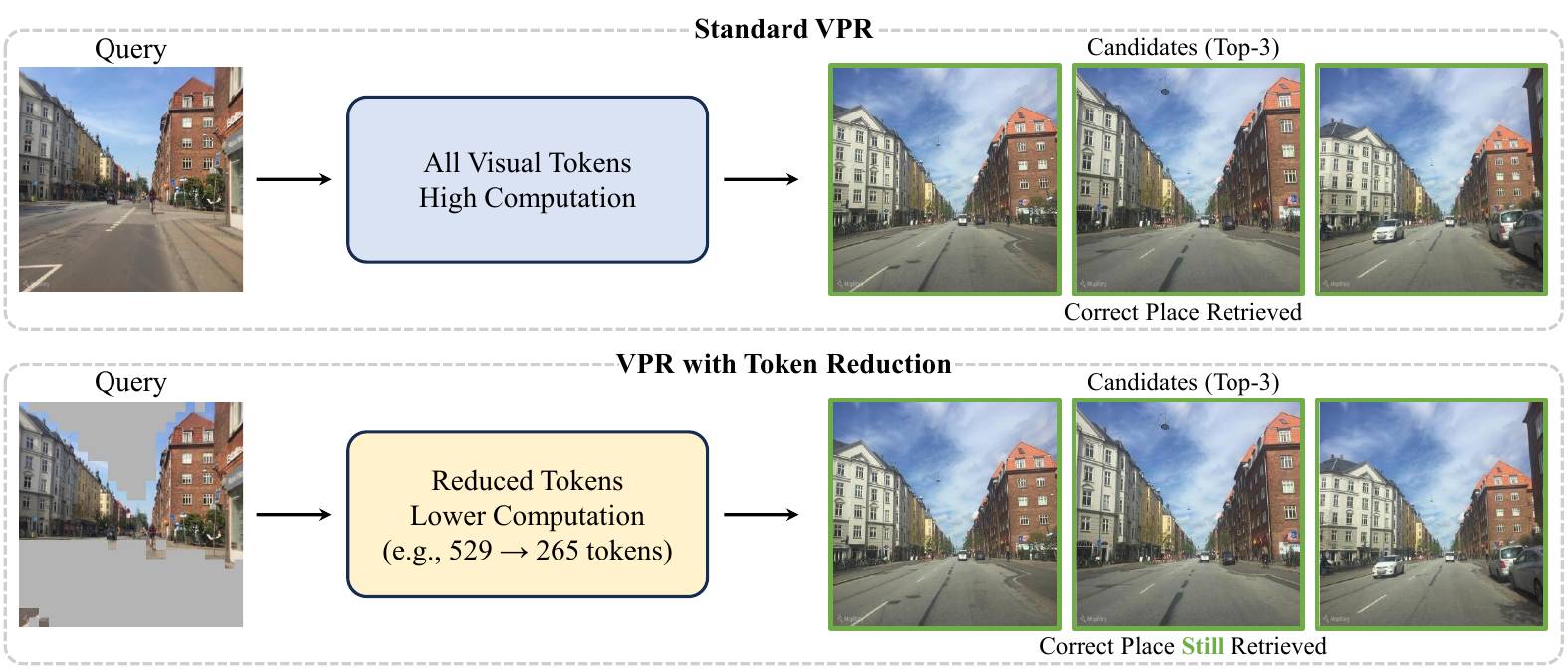}
    \caption{Illustration of token reduction for visual place recognition. In standard VPR, the query image is encoded into a sequence of visual tokens, which are then processed and aggregated into a global descriptor for retrieval. However, in token-reduced VPR, redundant tokens (denoted by gray patches ``\textcolor{gray}{\rule{1.2ex}{1.2ex}}'') corresponding to less informative regions are removed, while landmark-related tokens tend to be preserved. The correct place can still be retrieved despite using fewer tokens.}
    \label{fig:introduction}
\end{figure}

A key property of vision transformers is that their computational cost is highly dependent on the input sequence length. Most projection and feed-forward operations scale linearly with the number of tokens, while self-attention scales quadratically due to pairwise token interactions \cite{vit}. This suggests that reducing the number of visual tokens processed by the network can be a straightforward way to lower computational cost and accelerate inference. Additionally, prior VPR studies \cite{cricavpr, salad, boq, selavpr, edtformer, selavpr++} have shown that not all image regions contribute equally to place recognition. Static and distinctive structures such as buildings, signs, and landmarks are often critical for recognition, whereas less stable or less discriminative regions (e.g., sky, roads, pedestrians, and vehicles) tend to provide fewer reliable cues. This indicates that VPR may exhibit an inherent visual redundancy. As illustrated in Fig. \ref{fig:introduction}, many visual tokens in place images are possibly redundant for retrieval, while a subset of landmark-related tokens can preserve the discriminative cues required for correct matching. These observations naturally motivate the introduction of token reduction techniques \cite{survey_tr} into transformer-based VPR, with the goal of removing redundant tokens while preserving place-discriminative information. However, the behavior of token reduction in VPR remains largely unexplored. Existing token reduction methods have been extensively studied in image classification \cite{dynamicvit,ats,evit,tps,tome,pm_tmm} and multimodal large language models \cite{zs_prune,div_prune,graph_prune,llava_prumerge,agilepruner}, but VPR differs fundamentally from these settings. As a retrieval-oriented task, VPR requires generating globally discriminative descriptors whose reliability depends on both local place cues and their aggregated representation. Consequently, it remains unclear whether existing token reduction methods can effectively improve inference efficiency without compromising retrieval accuracy, and which design choices are critical for achieving a favorable accuracy-efficiency trade-off in VPR. This gap calls for a systematic and in-depth investigation of token reduction in visual place recognition.

In this paper, we present, to the best of our knowledge, the first systematic benchmark of token reduction for efficient visual place recognition. Centered around the question of ``Are all visual tokens necessary for VPR?'', our study aims to understand whether existing token reduction techniques can improve inference efficiency while preserving retrieval robustness. To this end, we adapt representative token pruning, token merging, and hybrid pruning-merging methods to transformer-based VPR models under a unified training-free framework. We evaluate these approaches across multiple state-of-the-art (SOTA) VPR architectures and benchmark datasets covering diverse environmental conditions. Beyond overall performance comparison, we conduct in-depth analyses from both algorithmic and deployment perspectives, including the effects of keep ratio, reduction position, reduction schedule, query/database reduction setting, backbone generalization, and input resolution. We further examine practical trade-offs among retrieval accuracy, computational complexity, latency, throughput, and qualitative token behavior, and validate the deployment benefit on an edge device. Through this benchmark, we provide a comprehensive understanding of token reduction in VPR and derive practical guidelines for future research on token-efficient visual retrieval.

The contributions of this work are summarized as follows:

\noindent \textbf{1)} To the best of our knowledge, this is the first work to systematically study token reduction for efficient visual place recognition. We build a unified, training-free, and plug-and-play benchmark framework that integrates seven representative token pruning, token merging, and hybrid pruning-merging methods into mainstream transformer-based VPR pipelines. This framework supports different backbones and aggregators, while adapting VPR aggregation modules to irregular token sequences after reduction, enabling fair and reproducible comparison across diverse models and benchmark datasets. Given the widespread use of large foundation models in recent VPR systems, this study is timely in addressing the key bottleneck caused by computational overhead. Moreover, our work is also significant for the real-time deployment of VPR models on resource-constrained edge devices.

\noindent \textbf{2)} Our experiments show that transformer-based VPR models contain substantial token redundancy, and many visual tokens can be removed or merged with only minor retrieval degradation. More importantly, we find that the effectiveness of token reduction is highly dependent on the reduction criterion and VPR setting. Through extensive analyses of keep ratio, reduction position, reduction schedule, query/database reduction setting, backbone generalization, and input resolution, we identify several key factors that govern the accuracy-efficiency trade-off of token reduction in retrieval-oriented VPR.

\noindent \textbf{3)} We further show that practical efficiency gains cannot be fully characterized by theoretical FLOPs alone. By measuring latency, throughput, and edge-device performance, we reveal that real acceleration depends on the reduction strategy, hardware platform, and inference mode. In particular, our edge-device experiments validate the deployment potential of token reduction in resource-constrained scenarios. Beyond efficiency measurements, our qualitative visualization and retrieval failure-case analysis further reveal the limitations of generic token reduction criteria, providing practical insights for future VPR-oriented token reduction methods.

\section{Related Work}
\label{sec:related work}

\subsection{Visual Place Recognition}
Early VPR methods \cite{rvpr, vpr_rs} mainly relied on hand-crafted local features such as SIFT \cite{sift} and SURF \cite{surf}, which were aggregated into global descriptors using classical schemes such as Bag of Words (BoW) \cite{bow} and VLAD \cite{vlad}. With the advent of deep learning, CNN-based approaches \cite{netvlad,gem,mixvpr,eigenplaces,cosplace,patch-netvlad,gcl} gradually became dominant, introducing both learned feature representations and trainable aggregation modules. Notable examples include NetVLAD \cite{netvlad}, which formulates differentiable VLAD encoding for end-to-end training, and GeM pooling \cite{gem}, which generalizes average pooling with a learnable parameter. More recently, transformer-based foundation models have been widely explored for VPR due to their strong representation ability \cite{cricavpr,salad,boq,selavpr,supervlad,emvp,edtformer,effovpr,fol,selavpr++}. Methods such as CricaVPR \cite{cricavpr}, SALAD \cite{salad}, BoQ \cite{boq}, SuperVLAD \cite{supervlad}, EffoVPR \cite{effovpr}, and SelaVPR++ \cite{selavpr++} build upon powerful pre-trained backbones, such as DINOv2 \cite{dinov2}, and design effective aggregation and fine-tuning strategies to obtain discriminative global descriptors. These foundation model-based methods have achieved remarkable accuracy gains and established new SOTA results on mainstream VPR benchmarks. Nevertheless, their reliance on large transformer backbones introduces substantially higher computational cost and inference latency than CNN-based methods. This accuracy-efficiency tension motivates the study of inference-efficient techniques for modern VPR systems.

\subsection{Token Reduction}
Token reduction, also referred to as token sparsification \cite{dynamicvit}, has been widely studied for accelerating transformer-based models by reducing the length of the input token sequence. Existing methods can be broadly divided into three paradigms: token pruning \cite{dynamicvit,ia-red2,ats,evo-vit,a-vit,evit,zs_prune}, token merging \cite{groupvit,tcformer,tome,tome4fsd}, and hybrid pruning-merging strategies \cite{tofu, llava_prumerge, pm_tmm}. Pruning-based methods discard redundant or less informative tokens according to certain importance criteria, such as learned prediction scores \cite{dynamicvit}, attention responses \cite{evit}, or token saliency. In contrast, merging-based methods reduce the sequence length by combining similar tokens rather than removing them, with ToMe \cite{tome} being a representative example that progressively merges tokens through bipartite soft matching. Hybrid methods further combine the two paradigms to exploit their complementary properties. For example, ToFu \cite{tofu} applies pruned merging in early layers and switches to average or norm-preserving merging in later layers, based on the observation that the suitability of pruning and merging varies across network depth. Beyond image classification, token reduction has recently been applied to multimodal large language models (MLLMs) \cite{fastv,fitprune,div_prune,graph_prune,llava_prumerge,agilepruner}, where visual tokens are typically compressed after the visual encoder before being fed into the large language model (LLM). In this setting, the language model usually dominates the computational cost, and token reduction mainly aims to reduce the burden of the subsequent LLM. However, VPR differs substantially from both image classification and MLLMs. A typical VPR model consists of a visual encoder followed by a lightweight aggregation module, making the encoder itself the primary computational bottleneck. \textit{Therefore, token reduction needs to be performed within the encoder to achieve meaningful acceleration, while preserving the place-discriminative cues required for generating reliable global descriptors.} These differences make the behavior and practical trade-offs of token reduction in VPR unclear, which greatly motivates the systematic benchmark conducted in this work.

\subsection{Efficient Visual Place Recognition}
Improving the efficiency of visual place recognition has been explored from several perspectives. Some methods focus on training efficiency \cite{gcl, edtformer, selavpr, selavpr++}, such as GCL \cite{gcl}, which introduces graded similarity labels to reduce costly hard-pair mining. Another line of work targets model-level compression for efficient inference, including network pruning \cite{mp4vpr} and weight quantization \cite{bnn4vpr1,bnn4vpr2,bnn4vpr3,bnn4vpr4}. Descriptor compression and quantization methods \cite{fq4vpr1,selavpr++} further reduce the storage and retrieval cost of global descriptors for large-scale place databases. More recently, asymmetric query processing \cite{asymvpr} reduces online query cost by using a lightweight query network. Closer to our topic, WeiToP \cite{weitop} explored VPR-oriented token pruning by learning aggregation-induced token importance from its proposed WeiAD aggregation framework. However, WeiToP couples token pruning with a specific learned aggregation design. \textit{Different from these method-oriented or model-level designs, our work systematically benchmarks token-level inference efficiency in a wide range of transformer-based VPR models under a unified training-free protocol, covering not only retrieval accuracy but also latency, throughput, and real edge-device deployment (Sec. \ref{sec:edge_device}). Instead of compressing parameters or final descriptors, token reduction shortens the visual token sequence inside the transformer encoder, directly targeting the dominant computational bottleneck of modern foundation-model-based VPR systems. This direction is complementary to existing efficient VPR techniques and can be combined with them for further efficiency gains.}

\section{Methodology}
\label{sec:methodology}

\subsection{Unified VPR Pipeline with Token Reduction}
\label{sec:pipeline}
Modern transformer-based VPR models commonly consist of a backbone and an aggregator. Given an input image $I \in \mathbb{R}^{H \times W \times 3}$, a patch embedding layer divides it into non-overlapping patches of size $p \times p$ and projects them into a sequence of $N=\frac{H}{p}\times\frac{W}{p}$ patch tokens. A learnable class token \texttt{[CLS]} is prepended, and positional embeddings are added to obtain the input sequence $\mathbf{X}_0 \in \mathbb{R}^{(N+1)\times D}$, where $D$ denotes the embedding dimension. The sequence is then processed by $L$ transformer blocks:
\begin{equation}
\mathbf{X}_l = \mathcal{B}_l(\mathbf{X}_{l-1}), \quad l=1,\dots,L ,
\end{equation}
where each block $\mathcal{B}_l$ consists of multi-head self-attention (MHSA) and feed-forward network (FFN) modules. After the final block, the output patch tokens are fed into an aggregation module, such as NetVLAD \cite{netvlad}, SALAD \cite{salad}, or BoQ \cite{boq}, to produce a compact global descriptor.

\begin{figure*}[t]
    \centering
    \includegraphics[width=1.0\linewidth]{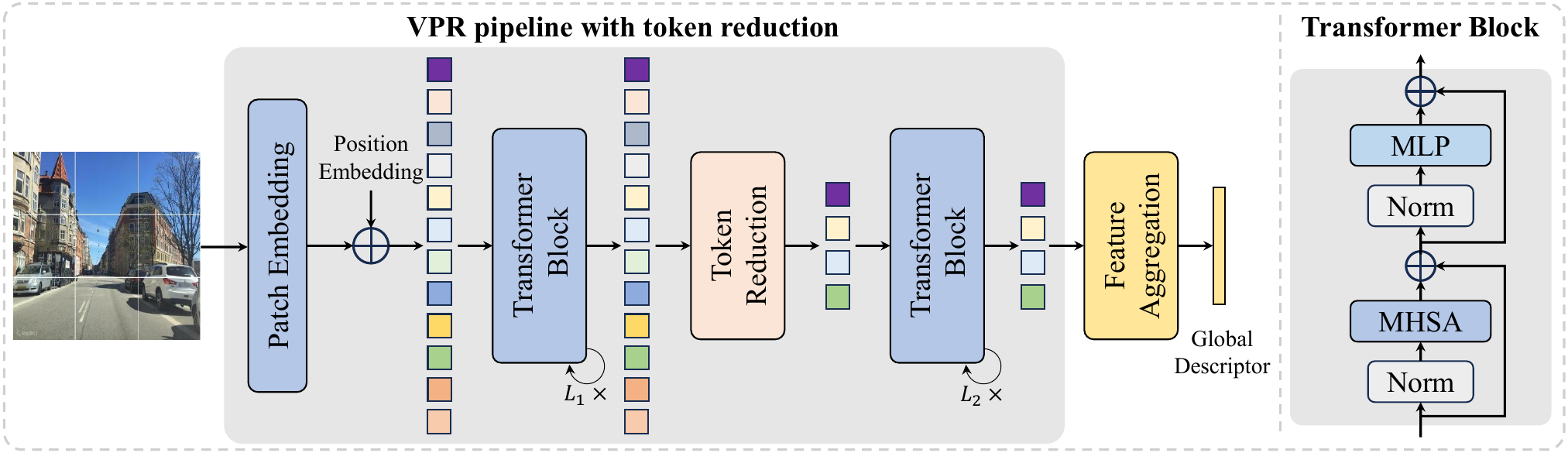}
    \caption{Overview of the VPR pipeline with token reduction. The input image is divided into patches and encoded as a token sequence. After passing through $L_1$ transformer blocks, a token reduction module removes or merges redundant tokens, and the reduced sequence is processed by the remaining $L_2$ blocks before being aggregated into a global descriptor for retrieval. The detailed structure of each transformer block is shown on the right.}
    \label{fig:pipeline}
\end{figure*}

To integrate token reduction into this pipeline, we insert a reduction operation $\mathcal{R}$ at an intermediate layer $s$. Given the token sequence $\mathbf{X}_s$, the reduction operation removes or merges patch tokens and produces a shorter sequence:
\begin{equation}
\widetilde{\mathbf{X}}_s = \mathcal{R}(\mathbf{X}_s), 
\quad 
\widetilde{\mathbf{X}}_s \in \mathbb{R}^{(K+1)\times D}, 
\quad K < N ,
\end{equation}
where $K$ denotes the number of retained or merged patch tokens. The reduced sequence is then processed by the remaining transformer blocks and subsequently aggregated into the final global descriptor. In this way, token reduction reduces the sequence length of all downstream layers, thereby lowering the computational cost of the visual encoder while preserving the standard VPR pipeline. Although Fig. \ref{fig:pipeline} illustrates token reduction at a single layer for clarity, the same formulation also supports progressive reduction across multiple layers, which is investigated in our experiments (Sec. \ref{sec:in-depth exploration and analysis}).

Notably, a practical issue is that some VPR aggregation modules are originally implemented on spatial feature maps with a fixed grid structure. After token reduction, the remaining tokens no longer necessarily form a regular spatial grid. To support arbitrary token lengths, we adapt such aggregators to operate on token sequences whenever necessary, without changing their aggregation principle. This enables different token reduction strategies to be evaluated under a unified and plug-and-play VPR pipeline.

\subsection{Token Reduction Design Space}
\label{sec:strategies}
In this study, we view token reduction as a general sequence transformation operation that converts the original token sequence into a shorter one. Given the patch-token sequence $\mathbf{X}=\{\mathbf{x}_i\}_{i=1}^{N}$ at a certain transformer layer, a token reduction operator $\mathcal{R}$ produces a reduced sequence:
\begin{equation}
    \widetilde{\mathbf{X}} = \mathcal{R}(\mathbf{X}; \rho),
    \quad
    \widetilde{\mathbf{X}}=\{\widetilde{\mathbf{x}}_j\}_{j=1}^{K},
    \quad
    K=\lfloor \rho N \rfloor,
\end{equation}
where $\rho \in (0,1]$ denotes the keep ratio and $K$ is the number of tokens after reduction. Under this formulation, existing token reduction strategies mainly differ in two aspects: how informative or redundant tokens are identified, and how the sequence length is reduced.

Specifically, token reduction can be formulated through two basic operators: token pruning and token merging. Token pruning reduces the sequence length by selecting a subset of tokens according to an importance function $f(\cdot)$:
\begin{equation}
    \widetilde{\mathbf{X}} =
    \{\mathbf{x}_i \mid i \in \operatorname{TopK}(f(\mathbf{x}_i, \mathbf{X}))\}.
\end{equation}
Different pruning methods mainly correspond to different definitions of $f(\cdot)$, such as random sampling, feature magnitude, attention response, etc. 

In contrast, token merging reduces the sequence length by combining similar tokens rather than discarding them:
\begin{equation}
    \widetilde{\mathbf{x}}_j = \sum_{i \in \mathcal{G}_j} w_i \mathbf{x}_i ,
\end{equation}
where $\mathcal{G}_j$ denotes a group of tokens to be merged and $w_i$ is the corresponding merging weight. Compared with pruning, merging preserves information from reduced tokens by fusing them into representative tokens. Hybrid pruning-merging methods further combine these two operators, typically applying pruning and merging at different layers or stages to exploit their complementary properties.

From this unified perspective, token reduction in VPR can be characterized along four key dimensions: \textit{1) the reduction operator, i.e., whether tokens are pruned, merged, or reduced through a hybrid combination; 2) the importance criterion, i.e., how informative or redundant tokens are identified; 3) the reduction position, i.e., at which transformer layer reduction is applied; and 4) the reduction schedule, i.e., whether reduction is performed once or progressively across multiple layers.} Our benchmark is designed to systematically evaluate these dimensions and their effects on VPR accuracy and efficiency.

\begin{figure*}[t]
    \centering
    \includegraphics[width=1.0\linewidth]{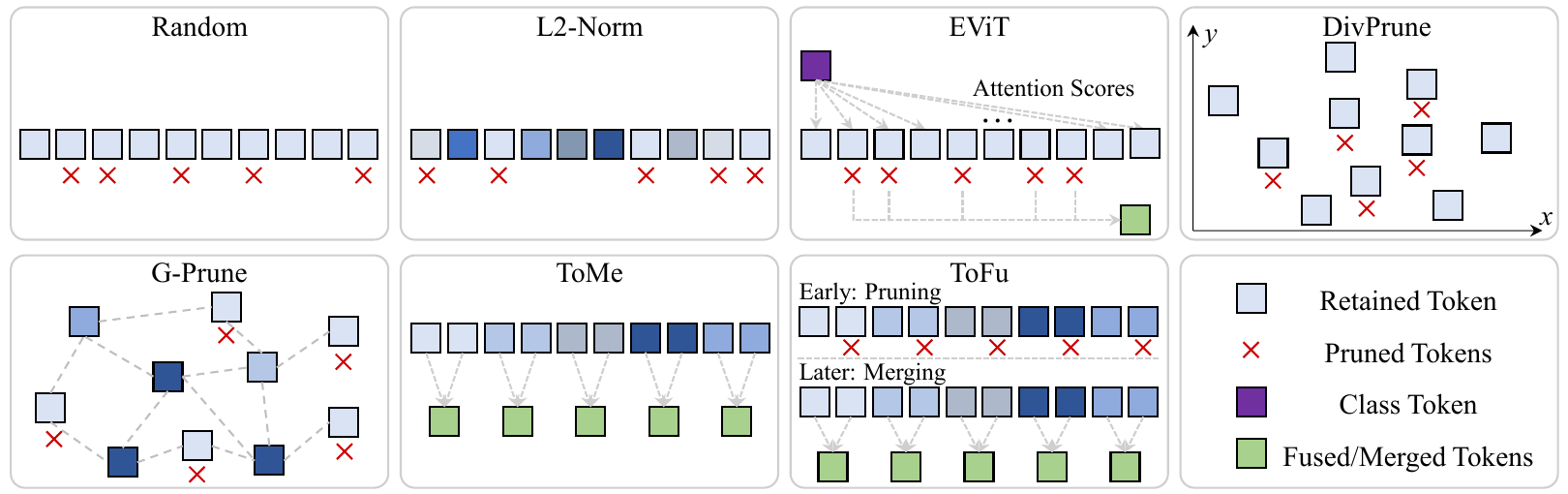}
    \caption{Illustration of the seven representative token reduction strategies evaluated in this work. Random, L2-Norm, EViT, DivPrune, and G-Prune belong to token pruning; ToMe belongs to token merging; and ToFu represents a hybrid pruning-merging strategy. The methods differ in how token importance or redundancy is estimated and how the token sequence is reduced.}
    \label{fig:tr_methods}
\end{figure*}

To cover a diverse design space, we evaluate a set of representative token reduction strategies, including simple baselines and more sophisticated methods spanning pruning, merging, and hybrid pruning-merging paradigms. All methods are applied in a training-free manner to well-trained VPR models, enabling a fair comparison of their intrinsic compatibility with VPR. An illustrative overview is shown in Fig. \ref{fig:tr_methods}.

\textbf{Random Selection.}
Random selection uniformly samples $K$ tokens from the $N$ patch tokens without any importance criterion. It serves as a simple lower-bound baseline for evaluating whether informed token selection is necessary.

\textbf{L2-Norm Pruning.}
L2-Norm pruning uses the feature magnitude as the token importance score. For each token $\mathbf{x}_i$, the score is computed as $s_i=\|\mathbf{x}_i\|_2$, and the top-$K$ tokens with the largest scores are retained. This represents a lightweight magnitude-based pruning strategy.

\textbf{EViT \cite{evit}.}
EViT estimates token importance using the attention scores between the class token \texttt{[CLS]} and patch tokens. Tokens receiving higher attention from the class token are retained as informative tokens. Following the original design, discarded tokens are fused into a new token through weighted averaging and forwarded to subsequent layers.

\textbf{DivPrune \cite{div_prune}.}
DivPrune selects tokens from a diversity perspective. It formulates token selection as a maximum diversity problem, encouraging the retained tokens to be well spread in the feature space. This criterion helps reduce redundancy among selected tokens and preserve representative visual information.

\textbf{G-Prune \cite{graph_prune}.}
G-Prune models tokens as graph nodes and estimates token importance through graph-based information propagation. Pairwise feature similarities are used to construct token relationships, and importance scores are propagated over the graph before selecting the top-$K$ tokens.

\textbf{ToMe \cite{tome}.}
ToMe represents token merging rather than token pruning. It reduces the sequence length by identifying similar token pairs through bipartite soft matching and merging them into representative tokens. In this way, ToMe preserves information from multiple tokens while reducing the number of tokens processed by subsequent layers.

\textbf{ToFu \cite{tofu}.} ToFu is a hybrid pruning-merging strategy. Similar to ToMe, it identifies highly similar token pairs through bipartite soft matching. Instead of always merging matched tokens, ToFu applies pruning in early layers and switches to average or MLERP merging in later layers, thereby combining token removal and merging across network depths.

Following the original designs of these methods, EViT, ToMe, and ToFu perform reduction between the MHSA and FFN modules, while the other pruning methods operate on the output of a complete transformer block. We preserve these design choices to avoid disadvantaging any method and to ensure a fair evaluation under their intended usage.

\subsection{Computational Analysis}
\label{sec:complexity}
We analyze the computational effect of token reduction to clarify why reducing the token sequence length can improve inference efficiency. In transformer-based VPR models, the transformer encoder usually dominates the overall computation, while the aggregation module accounts for a relatively small portion. Therefore, we focus on the computational cost of a standard transformer block with respect to the token length.

\begin{table}[t]
    \centering
    \caption{Computational complexity of a standard transformer encoder block before and after token reduction. $N$ and $K$ denote the number of tokens before and after Reduction, respectively, with $K<N$, and $D$ denotes the embedding dimension. Token reduction decreases linear operations by $K/N$ and self-attention operations by $(K/N)^2$. For simplicity, the class token is omitted in the complexity notation.}
    \label{table:complexity}
    \begin{tabular}{l c c c c}
        \toprule
        \multirow{2}{*}{Module} & \multirow{2}{*}{Operation} & Original FLOPs & After Reduction & Reduction \\
        &  & ($N$ tokens) & ($K$ tokens) & Ratio \\
        \midrule
        MHSA & QKV Projection & $3ND^2$ & $3KD^2$ & $K/N$ \\
        MHSA & Attention ($QK^\top$) & $N^2D$ & $K^2D$ & $(K/N)^2$ \\
        MHSA & Attention $\times V$ & $N^2D$ & $K^2D$ & $(K/N)^2$ \\
        MHSA & Out Projection & $ND^2$ & $KD^2$ & $K/N$ \\
        MLP  & FC Layer $\times 2$ & $8ND^2$ & $8KD^2$ & $K/N$ \\
        \midrule
        Total & -- & $12ND^2 + 2N^2D$ & $12KD^2 + 2K^2D$ & -- \\
        \bottomrule
    \end{tabular}
\end{table}

Given an input sequence with $N$ tokens and embedding dimension $D$, the dominant FLOPs of a transformer block can be approximated as:
\begin{equation}
\mathrm{FLOPs}(N) \approx 12ND^2 + 2N^2D ,
\end{equation}
where the first term mainly comes from linear projections and feed-forward layers, and the second term corresponds to the pairwise token interactions in self-attention. As summarized in Table \ref{table:complexity}, most projection and FFN operations scale linearly with the token count, while self-attention scales quadratically.

When token reduction is applied at block $l$ and reduces the number of patch tokens from $N$ to $K$, the computational cost of each subsequent transformer block becomes:
\begin{equation}
\mathrm{FLOPs}(K) \approx 12KD^2 + 2K^2D ,
\end{equation}
where $K<N$. Therefore, the linear components are reduced by $K/N$, while the self-attention component is reduced by $(K/N)^2$. Specifically, with DINOv2-Base ($D=768$, $N=529$ at an input resolution of $322 \times 322$), a keep ratio of 0.5 reduces the FLOPs of each subsequent block by approximately 52\%, with the self-attention component alone reduced by 75\%.

\subsection{Benchmark Protocol}
\label{sec:protocol}
To ensure a fair and reproducible comparison, all token reduction strategies are evaluated under a unified benchmark protocol. First, we apply token reduction in a training-free manner to the same set of well-trained VPR models. This allows us to isolate the effect of token reduction itself without introducing additional fine-tuning or architecture-specific optimization. Second, for each comparison, all strategies share the same input resolution, reduction position, and keep ratio. Third, by default, token reduction is applied only to query images, while database descriptors are extracted using the original full-token model. This setting reflects practical VPR deployment, where database descriptors can be precomputed in advance, whereas query images are typically encoded in real time under resource-constrained conditions \cite{asymvpr}. We adopt this query-only setting as our default protocol, while also investigating the symmetric setting, where token reduction is applied to both query and database images, in our in-depth exploration and analysis (Sec. \ref{sec:in-depth exploration and analysis}).

We evaluate token reduction from both retrieval and efficiency perspectives. Retrieval performance is measured by Recall@N, following standard VPR practice \cite{vprbenchmark}, while computational cost is measured by FLOPs. Since FLOPs reduction does not necessarily translate into proportional speedup, we further report inference latency and throughput to assess practical efficiency. Unless otherwise specified, latency and throughput measurements include the overhead introduced by token scoring, merging, similarity computation, or graph propagation, as such overhead directly affects real deployment. We also validate representative strategies on an edge device to examine their behavior under resource-constrained scenarios (Sec. \ref{sec:edge_device}).

\begin{table}[t]
    \centering
    \caption{Summary of the benchmark datasets in our experiments.}
    \label{tab:datasets}
    \begin{tabular}{ l l c c }
        \toprule
        \multirow{2}{*}{Dataset} & \multirow{2}{*}{Characteristics} & \multicolumn{2}{c}{Number} \\
        \cmidrule{3-4}
        & & Database & Queries \\
        \midrule
        Pitts30k \cite{pitts} & perspective shifts & 10,000 & 6,816 \\
        MSLS-val \cite{msls} & geographic diversity & 18,871 & 740 \\
        MSLS-challenge \cite{msls} & long-term span & 38,770 & 27,092 \\
        Tokyo24/7 \cite{tokyo247} & diurnal lighting variance & 75,984 & 315 \\
        Nordland \cite{nordland} & season changes & 27,592 & 27,592 \\
        \bottomrule
    \end{tabular}
\end{table}

\section{Experiments}
\label{sec:experiments}
\subsection{Datasets}
We evaluate token reduction on several widely used VPR benchmark datasets that cover diverse real-world challenges, as summarized in Table \ref{tab:datasets}. Pitts30k \cite{pitts} is collected from Google Street View and mainly evaluates robustness to viewpoint changes in urban scenes. MSLS \cite{msls} is a large-scale benchmark with images captured across urban, suburban, and natural environments, covering substantial geographic and temporal diversity. Tokyo24/7 \cite{tokyo247} focuses on severe day-night illumination changes, while Nordland \cite{nordland_original,nordland} consists of a wide variety of natural scenes with seasonal appearance variations captured from a fixed train-mounted camera. 

\subsection{Implementation Details}
To evaluate token reduction under mainstream transformer-based VPR settings, we construct three main model configurations by combining DINOv2-Base \cite{dinov2} with three widely used aggregation methods: NetVLAD \cite{netvlad}, SALAD \cite{salad}, and BoQ \cite{boq}. Unless otherwise specified, all main experiments are conducted on these DINOv2-Base models to provide a focused and comparable analysis across different reduction strategies. We train these models on the merged dataset proposed by SelaVPR++ \cite{selavpr++} with the multi-similarity loss~\cite{multiloss}, obtaining strong baselines representative of modern advanced transformer-based VPR systems. The input resolution is set to $224\times224$ during training and $322\times322$ during inference. We use the AdamW optimizer with an initial learning rate of $5\times10^{-5}$, which is halved every 3 epochs. The batch size is set to 120 places with 4 images per place, yielding 480 images per batch. Training is terminated via early stopping when the performance on MSLS-val shows no improvement for 8 consecutive epochs.

\begin{sidewaystable}
    \centering
    \caption{Comparison of different token reduction strategies. All methods are applied after the 6th block with a keep ratio of 0.5. Best results are highlighted in \textbf{bold}, and the second best are \underline{underlined}. ``No Red.'' denotes the no reduction baseline.}
    \label{tab:main_results}
    \setlength{\tabcolsep}{0.7mm}
    \begin{tabular}{c c | c c c | c c c c c c c c c c c c c c c}
    \toprule
    \multirow{2}{*}{Model} & \multirow{2}{*}{Method} & FLOPs$\downarrow$ & Latency$\downarrow$ & Throughput$\uparrow$ & \multicolumn{3}{c}{Pitts30k} & \multicolumn{3}{c}{MSLS-val} & \multicolumn{3}{c}{MSLS-challenge} & \multicolumn{3}{c}{Tokyo24/7} & \multicolumn{3}{c}{Nordland} \\
    \cmidrule(lr){6-8} \cmidrule(lr){9-11} \cmidrule(lr){12-14} \cmidrule(lr){15-17} \cmidrule(lr){18-20}
    & & (G) & (ms/image) & (images/s) & R@1 & R@5 & R@10 & R@1 & R@5 & R@10 & R@1 & R@5 & R@10 & R@1 & R@5 & R@10 & R@1 & R@5 & R@10 \\
    \midrule
    % ===== DINOv2-Base + NetVLAD =====
    \rowcolor{gray!20} \cellcolor{white}
    \multirow{7}{*}{\rotatebox{90}{\shortstack[c]{DINOv2-Base\\+ NetVLAD}}}
    & No Red.  & 50.91 & 13.6 & 121.7 & 93.9 & 97.3 & 98.0 & 94.7 & 97.6 & 98.0 & 76.5 & 86.2 & 88.0 & 97.1 & 98.1 & 98.4 & 97.0 & 99.1 & 99.4 \\
    & Random   & 37.68 & 11.5 & 166.3 & 92.8 & 97.1 & 98.1 & 92.8 & 96.9 & 97.3 & 73.9 & 84.2 & 86.4 & 95.2 & \textbf{97.8} & \textbf{98.1} & 92.8 & 97.5 & 98.4 \\
    & L2-Norm  & 37.68 & 11.6 & 166.0 & 93.2 & 97.1 & 98.1 & \underline{94.6} & \textbf{97.3} & 97.7 & 75.2 & \underline{85.3} & 87.3 & 94.3 & 97.1 & 97.5 & 94.9 & 98.2 & 98.9 \\
    & EViT     & 36.48 & 11.2 & 169.8 & \textbf{93.6} & \underline{97.2} & 98.1 & 93.9 & \textbf{97.3} & \textbf{98.0} & \underline{75.4} & \textbf{85.8} & \underline{87.6} & 94.3 & 97.1 & 97.5 & \underline{96.0} & \underline{98.7} & \underline{99.2} \\
    & DivPrune & 37.89 & 14.5 & 159.8 & \textbf{93.6} & \textbf{97.3} & \textbf{98.3} & 93.6 & \textbf{97.3} & \textbf{98.0} & \textbf{75.6} & \textbf{85.8} & \textbf{87.8} & 95.2 & 97.1 & \underline{97.8} & \textbf{96.2} & \textbf{98.8} & \textbf{99.3} \\
    & G-Prune  & 37.89 & 11.8 & 164.8 & 93.1 & 97.1 & 98.0 & \textbf{94.7} & \textbf{97.3} & \underline{97.8} & 75.3 & \underline{85.3} & 87.3 & 94.9 & 97.1 & 97.5 & 95.1 & 98.4 & 98.9 \\
    & ToMe     & 36.49 & 11.4 & 169.2 & \underline{93.3} & \textbf{97.3} & \textbf{98.3} & 94.3 & \underline{97.0} & \textbf{98.0} & 74.9 & 85.0 & 87.2 & \textbf{96.2} & \underline{97.5} & \textbf{98.1} & 95.3 & 98.4 & 99.0 \\
    & ToFu     & 36.49 & 11.3 & 169.5 & \underline{93.3} & \textbf{97.3} & \underline{98.2} & 94.2 & \underline{97.0} & 97.6 & 75.1 & 85.0 & 87.2 & \underline{95.9} & 97.1 & \textbf{98.1} & 95.2 & 98.4 & 99.0 \\
    \midrule
    \midrule
    % ===== DINOv2-Base + SALAD =====
    \rowcolor{gray!20} \cellcolor{white}
    \multirow{7}{*}{\rotatebox{90}{\shortstack[c]{DINOv2-Base\\+ SALAD}}}
    & No Red.  & 51.37 & 14.7 & 118.7 & 93.6 & 97.1 & 97.8 & 95.0 & 97.6 & 98.0 & 76.3 & 85.5 & 87.6 & 97.5 & 98.7 & 99.0 & 96.2 & 98.6 & 99.2 \\
    & Random   & 37.91 & 12.4 & 163.8 & 92.4 & 96.5 & 97.5 & 92.6 & 96.9 & 97.4 & 73.5 & 83.8 & 86.0 & 94.9 & \textbf{98.1} & \textbf{98.7} & 91.5 & 96.8 & 97.9 \\
    & L2-Norm  & 37.91 & 12.5 & 163.5 & 93.0 & \underline{96.9} & 97.7 & 93.9 & \underline{97.3} & 97.6 & 74.7 & 84.7 & \underline{87.1} & 94.6 & 96.8 & 97.8 & 94.2 & 97.7 & 98.4 \\
    & EViT     & 36.71 & 12.1 & 167.8 & 93.0 & \underline{96.9} & \underline{97.8} & 93.9 & 97.0 & \textbf{97.8} & 74.8 & \underline{84.8} & \underline{87.1} & 94.0 & \underline{97.5} & 98.1 & \textbf{95.1} & \textbf{98.3} & \textbf{99.0} \\
    & DivPrune & 38.12 & 15.1 & 158.2 & \textbf{93.6} & \textbf{97.0} & \underline{97.8} & \textbf{94.6} & \underline{97.3} & \textbf{97.8} & \textbf{75.2} & \textbf{85.1} & \textbf{87.3} & 94.9 & \textbf{98.1} & \underline{98.4} & \underline{94.9} & \underline{98.2} & \textbf{99.0} \\
    & G-Prune  & 38.12 & 12.6 & 163.2 & 93.2 & \underline{96.9} & \textbf{97.9} & 94.2 & \underline{97.3} & \textbf{97.8} & \underline{75.0} & \underline{84.8} & \underline{87.1} & 94.3 & 96.8 & 97.8 & 94.3 & 97.7 & 98.5 \\
    & ToMe     & 36.72 & 12.2 & 167.3 & 93.1 & \underline{96.9} & 97.7 & 94.2 & \textbf{97.4} & \underline{97.7} & 74.4 & 84.4 & 86.7 & \textbf{96.2} & \underline{97.5} & \underline{98.4} & 94.5 & 97.9 & \underline{98.7} \\
    & ToFu     & 36.72 & 12.2 & 167.4 & \underline{93.5} & \underline{96.9} & 97.7 & \underline{94.3} & \textbf{97.4} & \textbf{97.8} & 74.5 & 84.4 & 86.6 & \underline{95.2} & 97.5 & \underline{98.4} & 94.5 & 98.0 & \underline{98.7} \\
    \midrule
    \midrule
    % ===== DINOv2-Base + BoQ =====
    \rowcolor{gray!20} \cellcolor{white}
    \multirow{7}{*}{\rotatebox{90}{\shortstack[c]{DINOv2-Base\\+ BoQ}}}
    & No Red.  & 53.42 & 14.9 & 117.5 & 93.8 & 97.2 & 98.2 & 95.0 & 97.6 & 98.1 & 76.4 & 86.2 & 88.1 & 96.8 & 98.7 & 98.7 & 95.8 & 98.8 & 99.3 \\
    & Random   & 38.99 & 12.5 & 162.4 & 92.7 & 96.8 & 97.9 & 92.4 & 97.2 & 97.7 & 73.7 & 84.3 & 86.7 & \underline{96.5} & \textbf{98.4} & \textbf{98.7} & 91.4 & 97.0 & 98.2 \\
    & L2-Norm  & 38.99 & 12.6 & 162.2 & 93.0 & 97.0 & 97.9 & \underline{94.3} & \textbf{97.6} & 98.0 & 75.0 & 85.3 & 87.4 & 95.6 & \textbf{98.4} & \textbf{98.7} & 93.3 & 97.6 & 98.6 \\
    & EViT     & 37.80 & 12.2 & 166.8 & 93.2 & \underline{97.1} & \underline{98.1} & 93.9 & \textbf{97.6} & \textbf{98.2} & 74.7 & \underline{85.4} & \underline{87.6} & 94.0 & 97.8 & \underline{98.4} & \underline{94.5} & \textbf{98.4} & \underline{99.1} \\
    & DivPrune & 39.21 & 15.3 & 158.0 & \textbf{93.5} & \textbf{97.2} & \textbf{98.2} & 93.8 & \underline{97.4} & \underline{98.1} & \textbf{75.5} & \textbf{85.8} & \textbf{87.8} & 95.9 & \underline{98.1} & \underline{98.4} & \textbf{94.9} & \textbf{98.4} & \textbf{99.2} \\
    & G-Prune  & 39.21 & 12.7 & 161.9 & 93.2 & 97.0 & 97.9 & \textbf{94.5} & \textbf{97.6} & 98.0 & 75.0 & \underline{85.4} & 87.4 & 95.9 & \textbf{98.4} & \textbf{98.7} & 94.2 & 97.9 & 98.7 \\
    & ToMe     & 37.81 & 12.3 & 166.2 & \textbf{93.5} & \underline{97.1} & 98.0 & 93.9 & 97.2 & 97.8 & \underline{75.1} & 85.1 & 87.3 & 96.2 & \textbf{98.4} & \textbf{98.7} & 93.8 & \underline{98.1} & 98.9 \\
    & ToFu     & 37.81 & 12.3 & 166.4 & \underline{93.4} & \underline{97.1} & 97.9 & 93.5 & \underline{97.4} & 98.0 & 74.6 & 85.2 & 87.3 & \textbf{97.1} & \underline{98.1} & \textbf{98.7} & 93.9 & 98.0 & 98.8 \\
    \bottomrule
    \end{tabular}
\end{sidewaystable}

\subsection{Overall Comparison}
We first compare different token reduction strategies under a unified setting. Unless otherwise specified, all methods are applied to DINOv2-Base VPR models with a fixed keep ratio of 0.5, and token reduction is performed after the 6th transformer block. Table \ref{tab:main_results} reports the results across three representative aggregation methods, including NetVLAD, SALAD, and BoQ, on multiple VPR benchmark datasets. We report retrieval accuracy together with FLOPs, latency, and throughput to evaluate both recognition performance and inference efficiency. Specifically, throughput is measured at batch size 32 to assess batched inference efficiency, while latency is measured at batch size 1 to reflect real-time single query processing. Both metrics are measured on a single NVIDIA RTX A6000. We summarize the key observations as follows.

\begin{itemize}
    \item \textbf{Not all visual tokens are necessary for VPR.}
    Across all benchmark datasets, most token reduction strategies incur only minor retrieval degradation when 50\% of patch tokens are retained. Notably, this phenomenon is consistently observed across DINOv2-Base models with three different aggregators, suggesting that token redundancy is a general property of transformer-based VPR pipelines rather than an artifact of any specific VPR model. We further observe that R@5 and R@10 are generally more stable than R@1, indicating that token reduction often perturbs the fine-grained ranking order rather than completely removing the correct place from the retrieved candidate list (as shown in Fig. \ref{fig:retrieval_cases} in Sec. \ref{sec:qualitative experiments}). This is particularly meaningful for practical VPR systems that first retrieve a small set of candidates for subsequent verification or re-ranking. \textit{Processing all visual tokens is therefore not always necessary for reliable place recognition, making token reduction a viable direction for efficient VPR inference.}
    
    \item \textbf{The token selection criterion is critical.}
    Although all reduction methods use the same keep ratio, their retrieval performance differs noticeably. Random selection consistently causes larger degradation than informed strategies, especially on challenging datasets. For instance, when using the ``DINOv2-Base + BoQ'' model, Random reduces R@1 from 95.8\% to 91.4\% on Nordland, whereas EViT and DivPrune preserve much higher accuracy, achieving 94.5\% and 94.9\%, respectively. \textit{This indicates that VPR token reduction is not merely a matter of shortening the sequence length, but also preserving place-discriminative visual tokens is essential.} This observation is further supported by the token reduction visualizations in Fig. \ref{fig:token_reduction_visualization} in Sec. \ref{sec:qualitative experiments}.

    \item \textbf{Efficiency gains depend on the metric and reduction overhead.} All reduction strategies substantially reduce theoretical FLOPs, typically by about 26\%--29\%, and improve batched throughput on high-end GPUs from roughly 117--122 images/s to 158--170 images/s, corresponding to about 1.4$\times$ speedup. However, single query latency improves more modestly, usually by around 2.5 ms for lightweight strategies such as EViT and ToMe. This is because batch-size-one inference may not fully saturate high-end GPUs, so reducing token FLOPs does not necessarily yield proportional wall-clock speedup. Complex methods such as DivPrune further introduce additional token selection overhead, which can offset the saved transformer computation in single query inference. Nevertheless, such methods remain attractive for batched inference, where the overhead is amortized, and for edge devices, where transformer forwarding becomes the dominant bottleneck and token reduction brings more evident latency gains (Sec. \ref{sec:edge_device}). \textit{These results show that practical efficiency should be evaluated with deployment-aware metrics, rather than FLOPs alone.} In particular, batched throughput, single query latency, and edge-device performance jointly determine whether a reduction strategy is practically useful.

    \item \textbf{Reduction sensitivity varies across datasets.} The impact of token reduction is dataset-dependent. On Pitts30k, most informed strategies cause only minor R@1 degradation, indicating substantial redundancy in typical urban scenes. In contrast, Nordland is more sensitive to token removal, where random or less suitable reduction can cause larger drops. This may be because natural scenes and severe seasonal changes require preserving more subtle and globally distributed cues for reliable matching. \textit{Therefore, the optimal reduction strategy and keep ratio should be adapted to the deployment scenario and dataset characteristics.}
\end{itemize}

In summary, Table \ref{tab:main_results} demonstrates that token reduction is a feasible paradigm for efficient transformer-based VPR. Most informed reduction strategies can preserve competitive retrieval performance, indicating that not all visual tokens are necessary for VPR. Meanwhile, the results also reveal several non-trivial trade-offs: the choice of reduction criterion affects retrieval robustness, practical speedup depends on runtime overhead rather than FLOPs alone, and reduction sensitivity varies across datasets.

\begin{figure*}[t]
    \centering
    \includegraphics[width=1.0\linewidth]{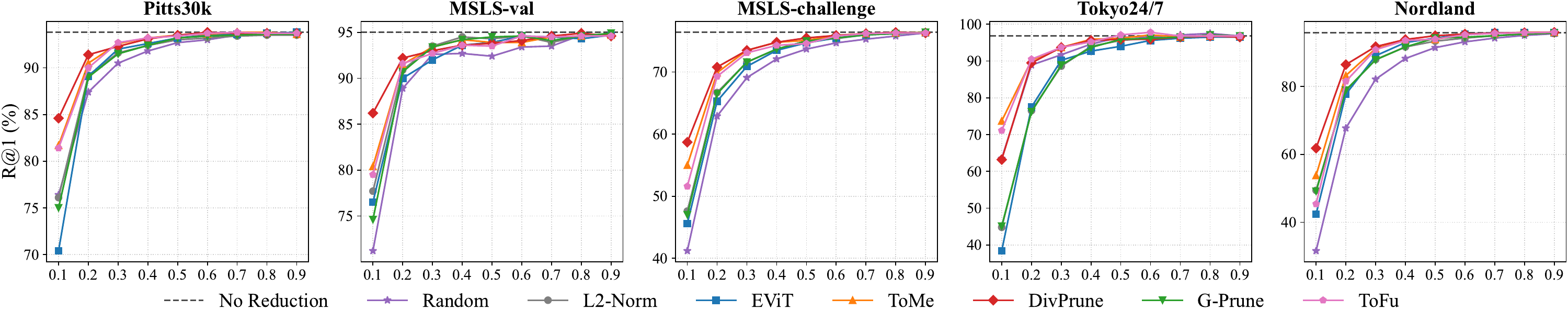}
    \caption{Effect of keep ratio on retrieval performance. Results are reported on ``DINOv2-Base + BoQ'' with token reduction applied after the 6th transformer block. The keep ratio varies from 0.1 to 0.9, and the dashed line denotes the no reduction baseline.}
    \label{fig:effect_of_keep_ratio}
\end{figure*}

\subsection{In-depth Exploration and Analysis}
\label{sec:in-depth exploration and analysis}
The overall comparison above is conducted under a fixed keep ratio and reduction position. \textit{To better understand the factors that govern the accuracy-efficiency trade-off of token reduction in VPR, we further analyze the effects of keep ratio, reduction position, reduction schedule, query/database reduction setting, backbone scales/types, and input resolution.} Unless otherwise specified, the following analyses are conducted on ``DINOv2-Base + BoQ'', which serves as a representative modern VPR configuration. The inference resolution is set to $322\times322$ for all experiments unless stated otherwise.

\textbf{1) Effect of Keep Ratio.}
We first study how the keep ratio affects retrieval performance. Fig. \ref{fig:effect_of_keep_ratio} shows the R@1 results of different token reduction strategies, where reduction is applied after the 6th transformer block, and the keep ratio varies from 0.1 to 0.9. The results reveal three key findings.
\begin{itemize}
    \item \textit{A moderate keep ratio is often sufficient for VPR.} Performance improves rapidly when the keep ratio increases from 0.1 to 0.3, but gradually saturates beyond 0.5. This suggests that retaining more tokens brings diminishing accuracy gains, and supports our default keep ratio of 0.5 used in the overall comparison.

    \item \textit{The reduction strategy becomes more important under aggressive reduction.} At low keep ratios, different methods show large performance gaps, indicating that preserving place-discriminative tokens is critical when only a small number of tokens can be retained. As the keep ratio increases, the gap among methods becomes smaller because most strategies can retain enough useful cues.

    \item \textit{Different datasets require different token budgets.} Pitts30k and MSLS-val recover quickly as the keep ratio increases, indicating that a relatively small subset of tokens can already preserve sufficient place cues in these scenarios. In contrast, Nordland improves more gradually and remains sensitive under low keep ratios, suggesting that natural scenes with severe seasonal changes require a larger token budget to preserve subtle or globally distributed cues for reliable matching.
\end{itemize}

\begin{figure*}[t]
    \centering
    \includegraphics[width=1.0\linewidth]{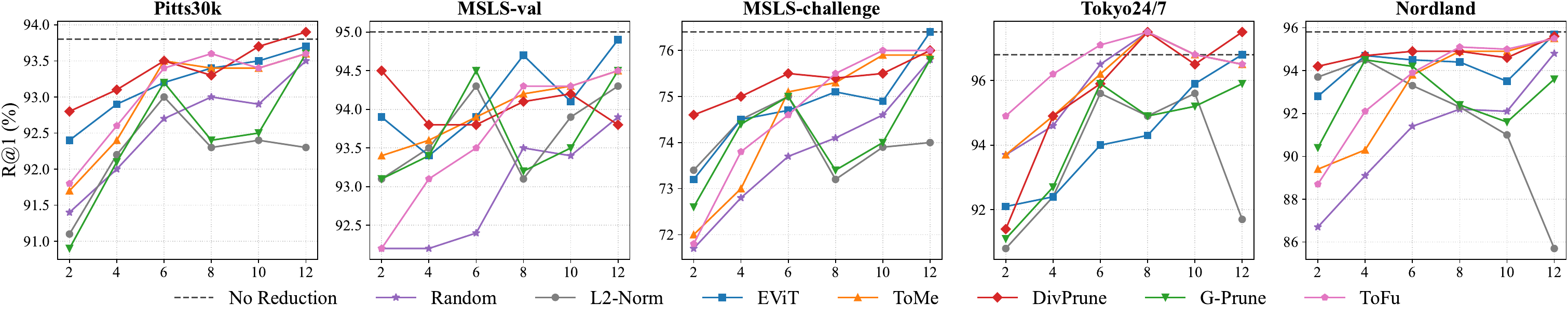}
    \caption{Effect of reduction position on retrieval performance. Results are reported on ``DINOv2-Base + BoQ'' with a fixed keep ratio of 0.5. Token reduction is applied after different transformer blocks, and the dashed line denotes the no reduction baseline.}
    \label{fig:effect_of_reduction_position}
\end{figure*}

\textbf{2) Effect of Reduction Position.}
We further study how the reduction position affects retrieval performance. Fig. \ref{fig:effect_of_reduction_position} reports R@1 when token reduction is applied at different transformer blocks with a fixed keep ratio of 0.5. Compared with the keep ratio analysis, the effect of reduction position is less strictly monotonic, since token representations evolve across transformer layers and different reduction criteria rely on different token statistics. Nevertheless, several trends can be observed.

\begin{itemize}
    \item \textit{Very early reduction is generally less stable.}
    Applying token reduction at early blocks often causes larger accuracy degradation, especially on challenging datasets such as MSLS-challenge and Nordland. This suggests that early-layer tokens may not yet encode sufficiently reliable place-discriminative information, making premature token removal risky.

    \item \textit{Middle-to-late layers provide a better accuracy-efficiency balance.} Reduction at middle or later blocks usually achieves more stable retrieval performance, because token representations have accumulated richer semantic and structural cues. However, applying reduction too late limits computational savings, since only a few downstream transformer blocks can benefit from the shortened sequence. Therefore, the reduction position should be selected by jointly considering retrieval accuracy and inference efficiency.

    \item \textit{The optimal reduction position is coupled with the reduction criterion.} Different strategies exhibit different sensitivities to the reduction position. For example, diversity-based or merging-based methods are generally more robust across positions, while simple magnitude-based selection may fluctuate on some datasets. This indicates that the reduction position and the token selection criterion should not be treated independently.
\end{itemize}

\begin{figure*}[t]
    \centering
    \includegraphics[width=1.0\linewidth]{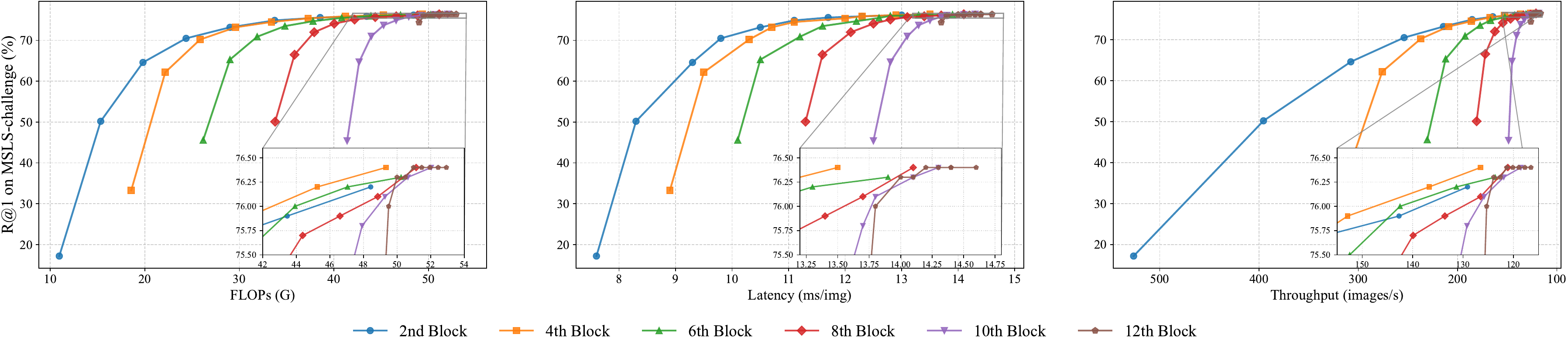}
    \caption{Accuracy-efficiency trade-off under different keep ratios and reduction positions. Each curve corresponds to a reduction position, and each point corresponds to a keep ratio from 0.1 to 0.9. The three subfigures compare R@1 with FLOPs, latency, and throughput, respectively.}
    \label{fig:accuracy_efficiency_tradeoff}
\end{figure*}

\textbf{3) Accuracy-Efficiency Trade-off.}
To jointly analyze the effects of keep ratio and reduction position, we plot the accuracy-efficiency trade-off on MSLS-challenge using the ``DINOv2-Base + BoQ'' model with EViT. For each reduction position, the keep ratio is varied from 0.1 to 0.9, producing trade-off curves between R@1 and FLOPs, latency, or throughput, as shown in Fig. \ref{fig:accuracy_efficiency_tradeoff}. \textit{The efficiency upper bound is mainly determined by how early tokens are reduced, rather than by how aggressively the keep ratio is lowered at later layers.} Earlier reduction allows more downstream transformer blocks to process shortened sequences, resulting in larger FLOPs reduction and more evident runtime benefits. In contrast, late-layer reduction provides limited acceleration even at low keep ratios, since only a few subsequent blocks can benefit from the reduced token length. However, early reduction also increases the risk of accuracy degradation, as shallow tokens may not yet encode sufficiently reliable place-discriminative cues. Therefore, the key challenge is to identify redundant tokens as early as possible without discarding retrieval-critical information. Overall, middle-layer reduction offers a practical compromise between accuracy preservation and efficiency gain, supporting our default setting in the overall comparison. Meanwhile, the trade-off curves suggest that stronger acceleration will require more reliable early-stage token selection mechanisms.

\begin{table}[t]
    \centering
    \caption{Effect of progressive reduction schedules on retrieval performance. We compare single-stage, three-stage, and per-block reduction using representative pruning and merging methods.}
    \label{tab:effect_of_progressive_reduction}
    \setlength{\tabcolsep}{1.2mm}
    \begin{tabular}{c c | c c | c c c c c c c c}
    \toprule
    \multirow{2}{*}{Method} & \multirow{2}{*}{Schedule} & FLOPs$\downarrow$ & Latency$\downarrow$ & \multicolumn{2}{c}{Pitts30k} & \multicolumn{2}{c}{MSLS-val} & \multicolumn{2}{c}{Tokyo24/7} & \multicolumn{2}{c}{Nordland} \\
    \cmidrule(lr){5-6} \cmidrule(lr){7-8} \cmidrule(lr){9-10} \cmidrule(lr){11-12}
    & & (G) & (ms/image) & R@1 & R@5 & R@1 & R@5 & R@1 & R@5 & R@1 & R@5 \\
    \midrule
    \multirow{3}{*}{EViT} & Single-stage & 37.80 & 12.2 & 93.2 & 97.1 & 93.9 & \textbf{97.6} & 94.0 & \textbf{97.8} & 94.5 & \textbf{98.4} \\
    & Three-stage & 37.08 & 12.1 & \textbf{93.4} & \textbf{97.2} & \textbf{94.5} & \textbf{97.6} & \textbf{94.9} & \textbf{97.8} & \textbf{95.0} & \textbf{98.4} \\
    & Per-block & 37.41 & 12.6 & 93.3 & 97.1 & 94.2 & 97.4 & 94.3 & \textbf{97.8} & 94.6 & \textbf{98.4} \\
    \midrule
    \multirow{3}{*}{ToMe} & Single-stage & 37.81 & 12.2 & 93.5 & 97.1 & 93.9 & 97.2 & 96.2 & \textbf{98.4} & 93.8 & 98.1 \\
    & Three-stage & 37.02 & 12.0 & 93.5 & \textbf{97.3} & 94.1 & 97.3 & 96.2 & \textbf{98.4} & 95.1 & \textbf{98.6} \\
    & Per-block & 37.25 & 13.0 & \textbf{93.6} & \textbf{97.3} & \textbf{94.3} & \textbf{97.4} & \textbf{96.5} & \textbf{98.4} & \textbf{95.2} & \textbf{98.6} \\
    \midrule
    \multirow{3}{*}{ToFu} & Single-stage & 37.81 & 12.2 & 93.4 & 97.1 & 93.5 & \textbf{97.4} & \textbf{97.1} & 98.1 & 93.9 & 98.0 \\
    & Three-stage & 37.02 & 12.1 & 93.5 & \textbf{97.3} & 94.3 & \textbf{97.4} & 96.5 & \textbf{98.4} & 95.0 & 98.5 \\
    & Per-block & 37.25 & 12.9 & \textbf{93.7} & \textbf{97.3} & \textbf{94.5} & \textbf{97.4} & 96.5 & \textbf{98.4} & \textbf{95.2} & \textbf{98.6} \\
    \bottomrule
    \end{tabular}
    \vspace{-0.5cm}
\end{table}

\textbf{4) Effect of Progressive Reduction.}
Several token reduction methods originally adopted progressive schedules. For example, EViT \cite{evit} reduces tokens at multiple transformer blocks, ToMe \cite{tome} performs merging across all blocks, and ToFu \cite{tofu} combines pruning and merging in different stages. We therefore investigate whether such schedules are also beneficial for VPR. Specifically, we compare single-stage, three-stage, and per-block reduction using representative pruning, merging, and hybrid pruning-merging methods. For a fair comparison, all schedules use the same final keep ratio of 0.5. Single-stage reduction is applied once after the 6th block, three-stage reduction is applied after the 4th, 6th, and 8th blocks, and per-block reduction gradually reduces tokens across all transformer blocks. As shown in Table \ref{tab:effect_of_progressive_reduction}, the results provide the following observations.
\begin{itemize}
    \item \textit{The optimal schedule depends on the reduction paradigm.} For pruning-based EViT, three-stage reduction achieves the best overall performance, suggesting that gradually removing tokens can alleviate the abrupt information loss caused by one-shot pruning. For merging-based ToMe, per-block reduction yields the best or comparable accuracy, which is consistent with its original design of progressively merging similar tokens across layers. For hybrid ToFu, progressive schedules also improve performance over single-stage reduction, indicating that combining pruning and merging can benefit from a staged reduction process.

    \item \textit{Progressive reduction improves FLOPs but does not always reduce latency.} Since token reduction starts earlier, progressive schedules generally reduce theoretical FLOPs. However, their runtime behavior differs. Three-stage reduction usually provides slightly better latency than single-stage reduction, while per-block reduction can be slower due to frequent reduction operations.

    \item \textit{Single-stage and three-stage reduction are both practical choices.} Single-stage reduction is simpler and better suited for unified comparison, whereas three-stage reduction can provide slightly better accuracy and efficiency when method-specific scheduling is allowed.
\end{itemize}

\textbf{5) Query-Only vs. Symmetric Reduction.}
Our default benchmark applies token reduction only to query images, since database descriptors are typically precomputed in practical VPR systems \cite{asymvpr}. However, in scenarios where database images need to be processed or updated frequently, such as dynamic map construction or resource-constrained database updating, it is also useful to examine whether token reduction can be applied to both query and database images. We therefore compare the default query-only setting with a symmetric setting, where the same reduction strategy is applied to both sides. To keep this analysis focused, we evaluate three representative lightweight strategies, including L2-Norm, EViT, and ToMe, which cover magnitude-based pruning, attention-based pruning, and token merging, respectively. As shown in Table \ref{tab:effect_of_query_only_pruning}, the comparison leads to the following observations.
\begin{itemize}
    \item \textit{Query-only reduction remains the practical default for real-time VPR.} Since database descriptors can usually be extracted offline with the full-token model, applying token reduction only to online queries directly reduces real-time inference cost while avoiding any modification to the database descriptor space.

    \item \textit{Symmetric reduction is also feasible when database descriptors must be updated online.} In most cases, applying token reduction to both query and database images introduces only limited additional degradation compared with query-only reduction. On MSLS-val and Tokyo24/7, symmetric reduction even achieves comparable or slightly better results for several methods, while the performance differences on Pitts30k and Nordland remain moderate.
    
    \item \textit{Consistent reduction on both sides does not severely distort the descriptor space.} The relatively small gap between query-only and symmetric reduction suggests that reduced-token descriptors can remain compatible when the same reduction strategy is applied consistently to both query and database images.
\end{itemize}

Overall, query-only reduction is better suited for static databases with precomputed descriptors, while symmetric reduction provides a feasible option for dynamic or resource-constrained database updating. This flexibility allows token reduction to adapt to different VPR deployment scenarios.

\begin{table}[t]
    \centering
    \caption{Comparison between query-only and symmetric token Reduction. ``Query-only'' denotes applying token reduction only to queries, while ``Symmetric'' denotes applying the same reduction strategy to both query and database images.}
    \label{tab:effect_of_query_only_pruning}
    \setlength{\tabcolsep}{2.7mm}
    \begin{tabular}{c c | c c c c c c c c c}
    \toprule
    \multirow{2}{*}{Method} & \multirow{2}{*}{Setting} & \multicolumn{2}{c}{Pitts30k} & \multicolumn{2}{c}{MSLS-val} & \multicolumn{2}{c}{Tokyo24/7} & \multicolumn{2}{c}{Nordland} \\
    \cmidrule(lr){3-4} \cmidrule(lr){5-6} \cmidrule(lr){7-8} \cmidrule(lr){9-10}
    & & R@1 & R@5 & R@1 & R@5 & R@1 & R@5 & R@1 & R@5 \\
    \midrule
    \multirow{2}{*}{L2-Norm} & Query-only & \textbf{93.0} & \textbf{97.0} & \textbf{94.3} & \textbf{97.6} & \textbf{95.6} & 98.4 & \textbf{93.3} & \textbf{97.6} \\
    & Symmetric & \textbf{93.0} & 96.9 & \textbf{94.3} & \textbf{97.6} & 94.6 & \textbf{99.0} & 92.6 & \textbf{97.6} \\
    \midrule
    \multirow{2}{*}{EViT} & Query-only & 93.2 & \textbf{97.1} & \textbf{93.9} & \textbf{97.6} & \textbf{94.0} & \textbf{97.8} & \textbf{94.5} & \textbf{98.4} \\
    & Symmetric & \textbf{93.3} & \textbf{97.1} & 93.6 & 97.4 & 93.7 & \textbf{97.8} & 92.9 & 97.8 \\
    \midrule
    \multirow{2}{*}{ToMe} & Query-only & \textbf{93.5} & \textbf{97.1} & 93.9 & 97.2 & \textbf{96.2} & \textbf{98.4} & \textbf{93.8} & \textbf{98.1} \\
    & Symmetric & 92.9 & 96.9 & \textbf{94.1} & \textbf{97.3} & 95.6 & 97.8 & 93.1 & 97.8 \\
    \bottomrule
    \end{tabular}
    \vspace{-0.5cm}
\end{table}

\begin{table}[t]
    \centering
    \caption{Token reduction results across different DINOv2 backbone scales. All methods are applied after the 6th block for DINOv2-Base and the 12th block for DINOv2-Large, with a keep ratio of 0.5.}
    \label{tab:backbone_scales}
    \setlength{\tabcolsep}{0.4mm}
    \begin{tabular}{c c | c c c | c c c c c}
    \toprule
    \multirow{2}{*}{Backbone} & \multirow{2}{*}{Method} & FLOPs$\downarrow$ & Latency$\downarrow$ & Throughput$\uparrow$ & \multirow{2}{*}{Pitts30k} & MSLS & MSLS & Tokyo & \multirow{2}{*}{Nordland} \\
    & & (G) & (ms/image) & (images/s) & & -val & -challenge & 24/7 \\
    \midrule
    \rowcolor{gray!20}
    \multirow{8}{*}{\rotatebox{90}{\shortstack[c]{DINOv2-Small}}}
    & No Red.  & 14.63 & 6.7 & 329.6 & 92.8 & 93.4 & 73.6 & 96.5 & 93.4 \\
    & Random   & 10.54 & 5.5 & 459.0 & 91.8 & 91.6 & 70.7 & 93.7 & 87.1 \\
    & L2-Norm  & 10.54 & 5.5 & 457.4 & 92.1 & 92.0 & 70.3 & 93.0 & 88.0 \\
    & EViT     & 10.24 & 5.4 & 465.2 & 92.2 & \textbf{93.2} & 72.0 & 93.0 & \textbf{91.7} \\
    & DivPrune & 10.64 & 9.7 & 426.2 & \textbf{92.5} & 93.0 & \textbf{72.5} & \textbf{94.9} & 91.4 \\
    & G-Prune  & 10.64 & 5.8 & 445.1 & 92.1 & 92.3 & 70.5 & 93.7 & 90.2 \\
    & ToMe     & 10.26 & 5.4 & 463.7 & \textbf{92.5} & 92.7 & 71.8 & \textbf{94.9} & 90.9 \\
    & ToFu     & 10.26 & 5.4 & 464.7 & \textbf{92.5} & 92.4 & 72.1 & 94.6 & 90.5 \\
    \midrule
    \rowcolor{gray!20}
    \multirow{8}{*}{\rotatebox{90}{\shortstack[c]{DINOv2-Large}}}
    & No Red.  & 178.76 & 35.7 & 37.8 & 94.7 & 95.8 & 78.8 & 98.1 & 97.6 \\
    & Random   & 131.41 & 28.5 & 52.9 & 93.2 & 93.1 & 75.9 & 97.1 & 93.9 \\
    & L2-Norm  & 131.41 & 28.6 & 52.7 & 93.7 & 94.7 & 77.0 & 96.8 & 96.5 \\
    & EViT     & 129.37 & 28.2 & 53.8 & 93.7 & 94.9 & 77.2 & 96.5 & 96.6 \\
    & DivPrune & 131.70 & 31.0 & 52.0 & \textbf{94.3} & \textbf{95.1} & \textbf{77.8} & 97.5 & \textbf{96.7} \\
    & G-Prune  & 131.70 & 28.7 & 52.2 & 93.7 & 94.7 & 77.0 & 96.5 & 96.5 \\
    & ToMe     & 129.28 & 28.3 & 53.5 & 93.8 & 94.5 & 76.8 & \textbf{97.8} & 96.0 \\
    & ToFu     & 129.28 & 28.3 & 53.6 & 93.9 & 94.3 & 77.0 & \textbf{97.8} & 96.1 \\
    \bottomrule
    \end{tabular}
    \vspace{-0.5cm}
\end{table}

\textbf{6) Backbone Generalization.}
To examine whether token reduction generalizes beyond the default DINOv2-Base backbone, we further evaluate different reduction strategies across both backbone scales and backbone types. Table \ref{tab:backbone_scales} reports results on DINOv2-Small and DINOv2-Large, while Table \ref{tab:backbone_types} evaluates two non-DINOv2 transformer backbones, including an ImageNet-pretrained ViT-B/16 and a CLIP-ViT-B/16 visual encoder. For the non-DINOv2 models, we follow prior practice \cite{vprbenchmark,image}. The last two blocks of ViT-B/16 are removed, and all remaining blocks are fine-tuned, while only the last six blocks are fine-tuned for CLIP-ViT-B/16. Both models are trained on the merged dataset \cite{selavpr++} with the multi-similarity loss at $224\times224$ resolution. We consistently use BoQ as the aggregator for all backbones. Several key insights are summarized as follows.

\begin{itemize}
    \item \textit{Token redundancy generalizes across DINOv2 backbone scales.} On both DINOv2-Small and DINOv2-Large, most informed reduction strategies preserve competitive R@1 while reducing FLOPs and improving throughput. This indicates that token redundancy is not specific to DINOv2-Base, but is also present in smaller and larger foundation backbones.

    \item \textit{Token reduction becomes particularly valuable for larger backbones.} On DINOv2-Large, FLOPs are reduced from 178.76G to around 129--132G, and throughput improves from 37.8 images/s to over 52 images/s for most lightweight strategies. Since larger backbones have heavier transformer computation, the relative overhead of complex reduction methods such as DivPrune becomes less dominant, making accuracy-oriented reduction strategies more practical on large models.

    \item \textit{Token redundancy also exists beyond DINOv2, but sensitivity varies across backbone types.} As shown in Table \ref{tab:backbone_types}, informed reduction strategies usually outperform random selection on both ViT-B/16 and CLIP-ViT-B/16, suggesting that meaningful token selection remains important across different pretraining paradigms. However, these backbones are more sensitive to token reduction than DINOv2, especially on challenging datasets. This may be due to less robust token representations for VPR compared with DINOv2, which makes generic reduction criteria less reliable. Another possible factor is the lower input resolution of $224\times224$, where fewer patch tokens are available and token removal is more likely to discard useful cues, as further examined in the input-resolution analysis below.
\end{itemize}

In short, these results confirm that token redundancy is a common property of transformer-based VPR models. Token reduction therefore provides an effective way to improve inference efficiency, especially for current VPR systems built upon large foundation backbones.

\begin{table}[t]
    \centering
    \caption{Token reduction results on non-DINOv2 transformer backbones. All methods are applied after the 5th block for ViT-B/16 and the 6th block for CLIP-ViT-B/16, with a keep ratio of 0.5. The inference resolution is $224\times224$.}
    \label{tab:backbone_types}
    \setlength{\tabcolsep}{0.4mm}
    \begin{tabular}{c c | c c c | c c c c c}
    \toprule
    \multirow{2}{*}{Backbone} & \multirow{2}{*}{Method} & FLOPs$\downarrow$ & Latency$\downarrow$ & Throughput$\uparrow$ & \multirow{2}{*}{Pitts30k} & MSLS & MSLS & Tokyo & \multirow{2}{*}{Nordland} \\
    & & (G) & (ms/image) & (images/s) & & -val & -challenge & 24/7 \\
    \midrule
    \rowcolor{gray!20}
    \multirow{8}{*}{\rotatebox{90}{\shortstack[c]{ViT-B/16}}}
    & No Red.  & 15.68 & 4.9 & 435.4 & 91.7 & 93.1 & 72.0 & 86.7 & 83.5 \\
    & Random   & 11.54 & 4.3 & 588.1 & 89.0 & 90.3 & 66.6 & 78.7 & 70.7 \\
    & L2-Norm  & 11.54 & 4.3 & 585.6 & 90.6 & \textbf{92.7} & 70.6 & 77.1 & \textbf{80.5} \\
    & EViT     & 11.13 & 4.2 & 601.4 & 90.7 & 92.4 & 70.0 & 75.2 & 79.9 \\
    & DivPrune & 11.57 & 7.1 & 569.9 & \textbf{91.3} & 92.0 & \textbf{70.6} & 81.9 & 79.9 \\
    & G-Prune  & 11.57 & 4.5 & 578.1 & 90.6 & 92.1 & 70.1 & 78.7 & 80.0 \\
    & ToMe     & 11.09 & 4.2 & 598.4 & 90.6 & 91.9 & 69.1 & \textbf{83.5} & 77.9 \\
    & ToFu     & 11.09 & 4.2 & 600.6 & 90.5 & 91.2 & 68.2 & 79.7 & 75.8 \\
    \midrule
    \rowcolor{gray!20}
    \multirow{8}{*}{\rotatebox{90}{\shortstack[c]{CLIP-ViT-B/16}}}
    & No Red.  & 18.59 & 5.6 & 341.4 & 91.5 & 93.5 & 72.0 & 89.8 & 80.6 \\
    & Random   & 13.71 & 5.0 & 463.0 & 88.6 & 90.1 & 66.6 & 83.2 & 67.2 \\
    & L2-Norm  & 13.71 & 5.0 & 461.9 & 90.1 & 91.4 & 69.3 & 81.9 & 72.3 \\
    & EViT     & 13.31 & 4.9 & 474.5 & 89.2 & 90.3 & 67.8 & 80.3 & 74.3 \\
    & DivPrune & 13.74 & 7.6 & 452.6 & \textbf{90.8} & 91.6 & \textbf{70.2} & \textbf{82.9} & \textbf{75.5} \\
    & G-Prune  & 13.74 & 5.2 & 457.6 & 89.9 & 90.7 & 67.2 & 72.4 & 74.5 \\
    & ToMe     & 13.26 & 4.9 & 472.1 & 90.1 & \textbf{92.3} & 69.7 & 87.3 & 75.3 \\
    & ToFu     & 13.26 & 4.9 & 473.9 & 89.2 & 90.9 & 67.7 & 85.1 & 70.1 \\
    \bottomrule
    \end{tabular}
    \vspace{-0.5cm}
\end{table}

\begin{table}[t]
    \centering
    \caption{Effect of input resolution on token reduction. We use ``DINOv2-Base + BoQ'' with EViT applied after the 2nd block and a keep ratio of 0.5. Accuracy denotes R@1 on Nordland, and red numbers indicate changes relative to the ``No Red.'' baseline at the same resolution.}
    \label{tab:effect_of_input_resolution}
    \setlength{\tabcolsep}{3.0mm}
    \begin{tabular}{c c c c c c}
    \toprule
    \multirow{2}{*}{Resolution} & \multirow{2}{*}{Method} & FLOPs$\downarrow$ & Latency$\downarrow$ & Throughput$\uparrow$ & \multirow{2}{*}{Accuracy} \\
    & & (G) & (ms/image) & (images/s) \\
    \midrule
    \multirow{2}{*}{224$\times$224} & No Red. & 24.85 & 7.9 & 265.9 & 92.1 \\
    & EViT & 13.90 \textcolor{red}{\footnotesize{(-44.1\%)}} & 6.2 \textcolor{red}{\footnotesize{(-21.5\%)}} & 451.9 \textcolor{red}{\footnotesize{(+70.0\%)}} & 84.6 \textcolor{red}{\footnotesize{(-7.5)}} \\
    \midrule
    \multirow{2}{*}{322$\times$322} & No Red. & 53.42 & 14.9 & 117.5 & 95.8 \\
    & EViT & 29.01 \textcolor{red}{\footnotesize{(-45.7\%)}} & 10.5 \textcolor{red}{\footnotesize{(-29.5\%)}} & 214.3 \textcolor{red}{\footnotesize{(+82.4\%)}} & 92.8 \textcolor{red}{\footnotesize{(-3.0)}} \\
    \midrule
    \multirow{2}{*}{448$\times$448} & No Red. & 112.24 & 32.6 & 44.2 & 97.4 \\
    & EViT & 59.16 \textcolor{red}{\footnotesize{(-47.3\%)}} & 19.8 \textcolor{red}{\footnotesize{(-39.3\%)}} & 96.0 \textcolor{red}{\footnotesize{(+117.2\%)}} & 96.3 \textcolor{red}{\footnotesize{(-1.1)}} \\
    \bottomrule
    \end{tabular}
    \vspace{-0.5cm}
\end{table}

\textbf{7) Effect of Input Resolution.}
We further investigate how token reduction behaves under different input resolutions. This analysis is conducted using ``DINOv2-Base + BoQ'', with EViT applied after the 2nd transformer block and a keep ratio of 0.5. We observe several consistent trends across resolutions in Table \ref{tab:effect_of_input_resolution}.
\begin{itemize}
    \item \textit{Token reduction brings consistent efficiency gains across resolutions.} EViT improves inference efficiency at all tested resolutions, indicating that token reduction is effective under different input sizes.

    \item \textit{The efficiency gain becomes larger at higher resolutions.} The latency reduction increases from 21.5\% at $224\times224$ to 39.3\% at $448\times448$, while the throughput improvement increases from 70.0\% to 117.2\%. This is because higher-resolution inputs produce more visual tokens, allowing token reduction to save more computation in subsequent transformer blocks.

    \item \textit{Higher-resolution inputs tolerate token reduction better.} The absolute R@1 drop decreases from 7.5 points at $224\times224$ to 1.5 points at $448\times448$. This suggests that high-resolution inputs contain more redundant tokens, allowing token reduction to remove a larger portion of computation while preserving sufficient place-discriminative cues.
\end{itemize}

Notably, token reduction is particularly valuable for high-resolution VPR scenarios, such as long-range landmark recognition or fine-grained structural matching, where simply lowering the input resolution may not be desirable.

\subsection{Edge Device Deployment}
\label{sec:edge_device}
Since VPR systems are often deployed on mobile robots or other resource-constrained platforms, FLOPs reduction and runtime measurements on high-end GPUs may not fully reflect practical deployment benefits. We therefore further evaluate whether token reduction can provide real latency improvement on an edge device. Experiments are conducted on an NVIDIA Jetson Xavier NX Developer Kit with 8GB unified memory, running JetPack 5.1.2. Latency is measured with batch size 1 to reflect real-time single query processing. We use the ``DINOv2-Base + BoQ'' model and evaluate two representative strategies: EViT as a lightweight attention-based method and DivPrune as a more complex diversity-based method.

\begin{table}[t]
    \centering
    \caption{Edge-device efficiency on NVIDIA Jetson Xavier NX. Speedup is computed relative to the ``No Red.'' baseline. Accuracy denotes R@1 on the Pitts30k dataset.}
    \label{tab:edge_latency}
    \setlength{\tabcolsep}{3.0mm}
    \begin{tabular}{l c c c c c c}
    \toprule
    \multirow{2}{*}{Method} & Keep  & Reduction & FLOPs $\downarrow$ & Latency $\downarrow$ & \multirow{2}{*}{Speedup $\uparrow$} & \multirow{2}{*}{Accuracy} \\
    & Ratio & Position & (G) & (ms/image) \\
    \midrule
    \rowcolor{gray!20} No Red. & -- & -- & 53.42 & 382.7 & 1.00$\times$ & 93.8 \\
    \midrule
    \multirow{9}{*}{EViT} & \multirow{3}{*}{0.5} & 6th & 37.80 & 285.3 & 1.34$\times$ & 93.2 \\
    & & 4th & 33.40 & 262.5 & 1.46$\times$ & 92.9 \\
    & & 2nd & 29.01 & 239.8 & 1.60$\times$ & 92.4 \\ 
    \cmidrule(lr){2-7}
    & \multirow{3}{*}{0.3} & 6th & 31.89 & 233.9 & 1.64$\times$ & 92.0 \\
    & & 4th & 25.85 & 197.2 & 1.94$\times$ & 90.4 \\
    & & 2nd & 19.81 & 160.5 & 2.38$\times$ & 88.7 \\
    \cmidrule(lr){2-7}
    & \multirow{3}{*}{0.1} & 6th & 26.18 & 200.5 & 1.91$\times$ & 70.4 \\
    & & 4th & 18.57 & 145.7 & 2.63$\times$ & 55.3 \\
    & & 2nd & 10.96 & 97.8 & 3.91$\times$ & 46.7 \\
    \midrule
    \multirow{9}{*}{DivPrune} & \multirow{3}{*}{0.5} & 6th & 39.21 & 311.1 & 1.23$\times$ & 93.5 \\
    & & 4th & 34.80 & 288.2 & 1.33$\times$ & 93.1 \\
    & & 2nd & 30.39 & 264.8 & 1.45$\times$ & 92.8 \\
    \cmidrule(lr){2-7}
    & \multirow{3}{*}{0.3} & 6th & 33.80 & 251.7 & 1.52$\times$ & 92.3 \\
    & & 4th & 27.75 & 215.2 & 1.78$\times$ & 91.8 \\
    & & 2nd & 21.69 & 177.9 & 2.15$\times$ & 89.7 \\
    \cmidrule(lr){2-7}
    & \multirow{3}{*}{0.1} & 6th & 28.60 & 211.0 & 1.81$\times$ & 84.6 \\
    & & 4th & 20.97 & 162.8 & 2.35$\times$ & 65.0 \\
    & & 2nd & 13.34 & 114.9 & 3.33$\times$ & 44.7 \\
    \bottomrule
    \end{tabular}
    \vspace{-0.5cm}
\end{table}

As shown in Table \ref{tab:edge_latency}, token reduction provides clear single query latency gains on the Jetson Xavier NX. \textit{This confirms that token reduction is especially beneficial on resource-constrained edge devices, where ViT forwarding becomes the dominant runtime bottleneck.} Under the moderate setting with a keep ratio of 0.5 after the 6th block, EViT reduces latency from 382.7 ms to 285.3 ms, achieving 1.34$\times$ speedup with only a small R@1 drop on Pitts30k. DivPrune follows a similar trend, achieving 1.23$\times$ speedup while maintaining competitive accuracy. These results suggest that moderate token reduction can already provide practical latency improvement without severely compromising retrieval performance.

More aggressive settings further reveal the acceleration potential of early reduction. For example, EViT with a keep ratio of 0.1 after the 2nd block reduces latency to 97.8 ms, corresponding to 3.91$\times$ speedup. However, this setting also causes severe accuracy degradation, indicating that extreme early reduction is not directly suitable without more reliable token selection. These results are consistent with our trade-off analysis. Earlier reduction can unlock much larger latency gains, but preserving retrieval accuracy requires reliable identification of redundant tokens at early layers. Therefore, practical edge deployment should choose the keep ratio and reduction position according to the required accuracy-latency trade-off, with lightweight strategies such as EViT being particularly suitable for real-time VPR.

\begin{figure*}[t]
    \centering
    \includegraphics[width=0.98\linewidth]{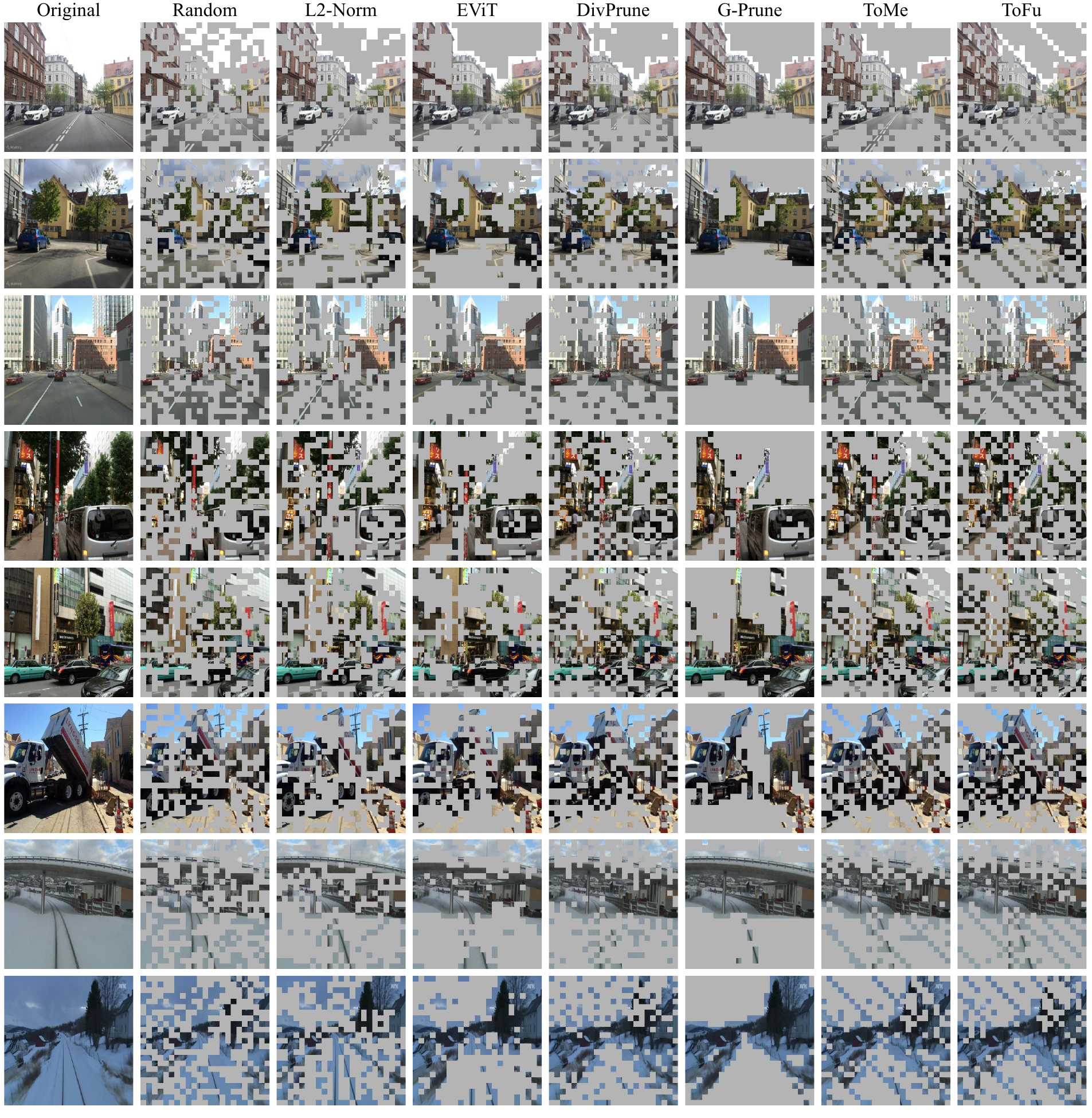}
    \caption{Visualization of token reduction results on representative VPR images. For pruning-based methods, masked patches denote discarded tokens, while for merging-based or hybrid methods, masked patches denote tokens merged or squeezed into representative tokens. The results show that existing generic reduction strategies can remove many low-information regions, but are not always aligned with VPR-specific place-discriminative cues.}
    \label{fig:token_reduction_visualization}
\end{figure*}

\begin{figure*}[t]
    \centering
    \includegraphics[width=0.98\linewidth]{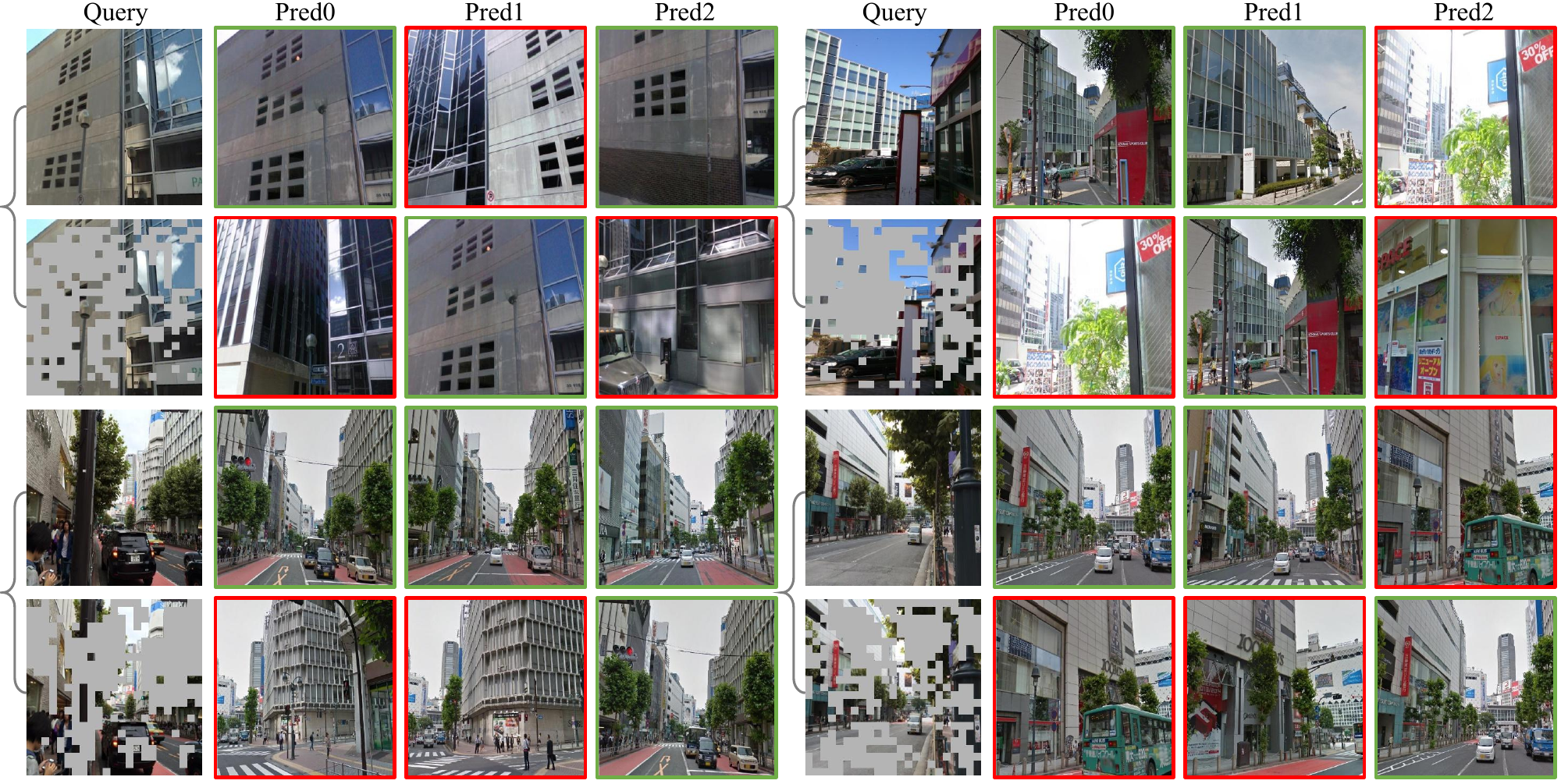}
    \caption{Failure cases of token reduction in VPR. For each query, we compare the top-3 retrieval results before and after token reduction. Green boxes indicate correct matches, while red boxes indicate incorrect retrieval results out of the predefined threshold. Token reduction may remove or compress subtle but place-discriminative cues, causing visually similar distractors to be ranked higher. In some cases, the correct place remains in the top-3 list, suggesting that token reduction often perturbs the fine-grained ranking order rather than completely destroying retrieval ability.}
    \label{fig:retrieval_cases}
\end{figure*}

\subsection{Qualitative Analysis}
\label{sec:qualitative experiments}
While the quantitative results demonstrate the accuracy-efficiency trade-offs of token reduction, they do not fully explain how different strategies behave at the token level. We therefore conduct qualitative analysis from two perspectives. First, we visualize the retained, removed, or merged tokens to understand which image regions are compressed by different methods. Second, we analyze failure cases by comparing the top-3 retrieval results before and after token reduction. These analyses provide further insights into the strengths and limitations of existing token reduction strategies for VPR.

\textbf{1) Token Retention Visualization.}
Fig. \ref{fig:token_reduction_visualization} visualizes the token reduction results of different strategies on representative VPR images. For pruning-based methods, masked patches denote discarded tokens. For merging-based or hybrid methods, masked patches indicate tokens that are merged or squeezed into representative tokens rather than directly removed. Several interesting patterns can be observed. 
\begin{itemize}
    \item \textit{Most informed strategies tend to compress visually redundant regions.} These regions include sky, roads, snow-covered ground, and other large homogeneous background areas. This supports our quantitative finding that VPR images contain substantial token redundancy, and that many tokens can be compressed without severely affecting retrieval performance.

    \item \textit{Similar retrieval performance can be achieved with different retained-token patterns.} Different strategies produce noticeably different token layouts, even when their retrieval accuracy is close. This suggests that VPR does not necessarily require a unique subset of tokens, as long as sufficient place-discriminative cues are preserved.

    \item \textit{Generic token importance is not always aligned with VPR-specific importance.} Existing methods are mainly guided by general criteria such as feature magnitude, attention response, diversity, graph similarity, or token similarity. They do not explicitly distinguish VPR-relevant static landmarks from less reliable dynamic foreground objects. As a result, some methods may preserve pedestrians, vehicles, or other visually salient but unstable objects, as shown in rows 2 and 4--6 of Fig. \ref{fig:token_reduction_visualization}, while occasionally removing parts of buildings, facades, signs, or structural layouts that are important for place recognition.
\end{itemize}

These observations suggest that future VPR-oriented token reduction methods should consider not only visual saliency or redundancy, but also the stability and discriminativeness of place-related cues.

\textbf{2) Failure Case Analysis.}
We further analyze several representative failure cases by comparing the top-3 retrieval results before and after token reduction. As shown in Fig. \ref{fig:retrieval_cases}, token reduction may change the retrieval ranking when subtle but place-discriminative cues are removed or excessively compressed. We summarize two main findings as follows.
\begin{itemize}
    \item \textit{Failures often arise from losing subtle structural cues rather than removing obviously salient regions.} In these cases, the global descriptor becomes less discriminative and is shifted toward visually similar distractors, leading to incorrect top-1 retrieval. A closer inspection shows that failures often occur when important structural cues, such as building facades, window layouts, road boundaries, or distinctive local textures, are partially removed or merged. Although these regions may not always be visually salient, they can be crucial for distinguishing between highly similar places.
    \item \textit{Token reduction often perturbs fine-grained ranking rather than completely destroying retrieval ability.} The correct place is sometimes still included in the top-3 results after token reduction, even when the top-1 prediction becomes incorrect. This observation is consistent with our quantitative results, where R@5 and R@10 are generally more stable than R@1. It also suggests that token-reduced descriptors may still be useful for fast preliminary candidate retrieval, followed by verification or re-ranking.
\end{itemize}

In summary, the qualitative results reveal both the promise and limitation of current token reduction strategies. They can effectively compress redundant visual regions, but still lack explicit awareness of stable, place-discriminative cues required by VPR.

\section{Discussion}
\textbf{1) Token redundancy exists, but VPR-specific importance is needed.}
Our experiments show that transformer-based VPR models contain substantial token redundancy, and a wide range of informed reduction strategies can remove a large portion of tokens while preserving competitive retrieval performance. However, random selection consistently leads to larger degradation, indicating that token selection criteria matter. Furthermore, our qualitative analysis reveals that generic reduction criteria may preserve visually salient but transient objects, such as vehicles and pedestrians, while removing subtle but place-discriminative structures, such as facades, signs, and layout cues. \textit{This misalignment reflects the unique nature of VPR, where reliable recognition depends on cues that remain stable across temporal, weather, and viewpoint variations, rather than on visually prominent regions in a single image.} Generic token importance, which is typically defined by feature magnitude, attention response, diversity, or local saliency, does not explicitly capture this notion of long-term stability. Future methods may therefore benefit from explicitly modeling stable landmarks, suppressing transient objects, preserving scene layout, and maintaining retrieval consistency across appearance variations.

\textbf{2) Practical acceleration is deployment-dependent and particularly valuable for VPR.}
Token reduction substantially decreases theoretical FLOPs, but its practical benefit depends on hardware and inference scenarios. On high-end GPUs, single query latency improvement can be modest, while batched throughput is significantly improved. On resource-constrained edge devices, reducing token sequence length leads to substantial single-query latency speedup because ViT forwarding becomes the dominant bottleneck. \textit{This deployment behavior is particularly relevant to VPR, where database descriptors are usually precomputed offline and online queries must be encoded in real time.} Moreover, our input-resolution analysis shows that the relative efficiency gain of token reduction grows with input resolution, further amplifying its value in high-resolution VPR scenarios.

\textbf{3) Reliable early-stage reduction is a promising direction.}
Our trade-off analysis shows that the reduction position strongly affects the efficiency upper bound, since earlier reduction allows more downstream blocks to benefit from the shortened sequence. However, early reduction is risky because shallow layers primarily encode low-level features and have not yet formed sufficient place-discriminative abstractions. Recent VPR-oriented pruning work, such as WeiToP \cite{weitop}, has explored early-stage token pruning, yet maintaining retrieval accuracy under aggressive early reduction remains challenging. \textit{A promising route is to transfer place-aware signals from deeper layers or retrieval consistency into lightweight early-layer selectors.} Possible directions include distilling deep-layer importance into shallow predictors, leveraging spatial priors to suppress clearly homogeneous regions before semantic features mature, or progressively refining token selection across multiple layers.

\textbf{4) Limitations and open questions.}
Our benchmark intentionally focuses on inference-time, training-free token reduction to ensure that the evaluated strategies can be directly integrated into existing VPR models without retraining. As a result, training-aware approaches that jointly optimize token reduction with VPR objectives, as explored by DynamicViT \cite{dynamicvit} in image classification, are not covered in this work. \textit{Whether VPR-specific training-aware reduction can further improve the accuracy-efficiency trade-off remains an important open question.} We hope our benchmark provides a useful foundation for addressing this question and inspires further research toward efficient VPR systems.

\section{Conclusion}
\label{sec:conclusion}
In this paper, we presented a systematic benchmark of token reduction for efficient visual place recognition. Under a unified transformer-based VPR pipeline, we evaluated representative pruning, merging, and hybrid reduction strategies across diverse models, datasets, and deployment settings. Our experiments show that transformer-based VPR models contain substantial token redundancy, and informed reduction strategies can remove or compress many visual tokens while maintaining competitive retrieval performance. We further demonstrate that practical acceleration should be assessed beyond FLOPs, since real latency and throughput are affected by reduction overhead, hardware characteristics, and deployment scenarios. Qualitative analysis reveals that generic token reduction methods are not explicitly aligned with VPR-specific importance, especially in preserving stable place-discriminative cues. Overall, this benchmark provides a comprehensive empirical foundation for understanding token reduction in VPR, and we hope our empirical study will inspire future research on VPR-oriented token reduction and efficient visual retrieval.

\section*{Declarations}
\textbf{Data Availability} The datasets used in this study are publicly available from their respective sources. For training, we follow the merged dataset construction proposed by SelaVPR++ \cite{selavpr++}, which consists of GSV-Cities, MSLS-train, Pitts30k-train, and a subset of SF-XL. The detailed construction protocol is available at \url{https://github.com/Lu-Feng/SelaVPRplusplus}. For evaluation, the test datasets follow the data organization protocol of the deep visual geo-localization benchmark \cite{vprbenchmark} and can be prepared using \url{https://github.com/gmberton/VPR-datasets-downloader}. All datasets used in this study are clearly cited in the manuscript.

\bibliography{main}

\end{document}